\documentclass[11pt]{article}

\usepackage[]{ACL2023}

\usepackage{times}
\usepackage{latexsym}

\usepackage[T1]{fontenc}

\usepackage[utf8]{inputenc}

\usepackage{microtype}
\usepackage{threeparttable}
\usepackage{tablefootnote}
\usepackage{booktabs}
\usepackage{microtype}
\usepackage{amsmath}
\usepackage{amsfonts} 
\usepackage{multirow}
\usepackage{xcolor}
\usepackage{tabularx}
\usepackage{makecell}
\usepackage{inconsolata}
\usepackage{graphicx}
\usepackage{tikz}
\usepackage{tikz-qtree}
\usetikzlibrary{trees}
\usepackage{forest}
\usetikzlibrary{fit}

\definecolor{color1}{RGB}{232,232,232}
\definecolor{color2}{RGB}{255,240,245}
\definecolor{color3}{RGB}{224,255,255}
\definecolor{color4}{RGB}{255, 235, 178}

\title{Self-assessment, Exhibition, and Recognition: a Review of Personality in Large Language Models}

\author{Zhiyuan Wen\textsuperscript{1}, Yu Yang\textsuperscript{1}, Jiannong Cao\textsuperscript{1}, Haoming Sun\textsuperscript{1}, Ruosong Yang\textsuperscript{1}, Shuaiqi Liu\textsuperscript{2}\\
  \textsuperscript{1}The Hong Kong Polytechnic University, Kowloon, Hong Kong, China\\
  \textsuperscript{2}Huawei Technologies Co., Ltd, Shenzhen, China\\
  \texttt{\{zyuanwen, cs-yu.yang, jiannong.cao\} @poly.edu.hk} \\
  \texttt{haoming.sun@connect.polyu.hk} \\
  \texttt{rsong.yang@polyu.edu.hk} \\
  \texttt{liushuaiqi@huawei.com}\\
}
\begin{document}
\maketitle

\begin{abstract}

As large language models (LLMs) appear to behave increasingly human-like in text-based interactions, more and more researchers become interested in investigating personality in LLMs. However, the diversity of psychological personality research and the rapid development of LLMs have led to a broad yet fragmented landscape of studies in this interdisciplinary field. Extensive studies across different research focuses, different personality psychometrics, and different LLMs make it challenging to have a holistic overview and further pose difficulties in applying findings to real-world applications. 
In this paper, we present a comprehensive review by categorizing current studies into three research problems: self-assessment, exhibition, and recognition, based on the intrinsic characteristics and external manifestations of personality in LLMs. For each problem, we provide a thorough analysis and conduct in-depth comparisons of their corresponding solutions. Besides, we summarize research findings and open challenges from current studies and further discuss their underlying causes. We also collect extensive publicly available resources to facilitate interested researchers and developers. Lastly, we discuss the potential future research directions and application scenarios. Our paper is the first comprehensive survey of up-to-date literature on personality in LLMs. By presenting a clear taxonomy, in-depth analysis, promising future directions, and extensive resource collections, we aim to provide a better understanding and facilitate further advancements in this emerging field.

\end{abstract}

\section{Introduction}

Large Language Models (LLMs) have exhibited impressive language comprehension and generation capabilities, enabling them to conduct coherent, human-like conversations with users. These remarkable progress have led to a wide range of applications \cite{chen2023large, zheng2023building, he2023psychological} and also ignited a growing interest in exploring the personality in LLMs.

Personality is described as the enduring characteristics that shape an individual's thoughts, emotions, and behaviors \cite{mischel2007introduction}. In the context of LLMs, researchers are curious about whether LLMs have intrinsic personality traits or how well can LLMs handle personality-related tasks in interaction.
These investigations facilitate understanding the psychological portrayal of LLMs \cite{huang2023chatgpt} and further constructing AI systems that are more transparent, safe, and trustworthy \cite{safdari2023personality}.

In light of this, numerous studies have emerged in this interdisciplinary field over the past two years, as shown in Appendix \ref{Overview}. However, the diversity of psychological personality research \cite{hodo2006kaplan} and the rapid development of LLMs make it difficult to not only obtain a comprehensive overview of this research area but also compare different methods, derive general conclusions, and apply findings to real-world applications. Specifically, current studies exhibit a hodgepodge in:

\begin{itemize}
\item \textbf{Research Focuses}: The topic of personality in LLMs encompasses various aspects, \textit{e.g.}, LLMs' personality assessment, or LLMs' awareness of users' personalities. Despite this breadth, most studies are only interested in particular aspects. 
\item \textbf{Psychometrics}: Different studies focus on different personality models (\textit{e.g.}, Big-five \cite{digman1990personality} and the Myers-Briggs Type Indicator (MBTI; \citet{Myers_1962}). Even for the same personality model, researchers may also adopt different psychometrics in their works.

\item \textbf{Investigated LLMs}: Over the past two years, numerous LLMs have been released. Despite a common focus on personality in LLMs, different researchers investigate different LLMs.
\end{itemize}

\begin{figure*}
\scriptsize
\begin{forest}
  for tree={
  	edge path={
      \noexpand\path [draw, \forestoption{edge}] 
      (!u.parent anchor) -- +(1em,0em) |- (.child anchor);
    },
    parent anchor=east,
    child anchor=west,
    grow'=0,
    align=center,
    l sep+=10pt,
    draw,
    rounded corners,
    level 1+/.style={level distance=25pt},
    level distance=50pt, 
    font = {\scriptsize}
  },
  [Personality \\ in LLMs, fill=color1!40, rounded corners, minimum width=2cm,
    [Self-assessment \\ (Section \ref{Self-assessment}), fill=color2!40, minimum width=2cm, rounded corners, 
    	[\textbf{Likert-scale Questionnaires} \\ 
     \cite{caron2022identifying};\cite{jiang2023evaluating};\cite{song2023large};\cite{miotto2022gpt3}; \\ \cite{li2023does};\cite{lacava2024open};\cite{sorokovikova2024llms};\cite{huang2023chatgpt}; \\ 
     \cite{safdari2023personality}; \cite{pan2023llms};\cite{yu2023personality};\cite{ai2024cognition} \\ \cite{barua2024psychology};\cite{stockli2024personification};\cite{salecha2024large};\cite{pellert2024ai}; \\ 
      \cite{bodroza2023personality};\cite{huang2023chatgptMBTI};\cite{huang2023revisiting};\cite{romero2023gpt};\\
      \cite{dorner2023personality};\cite{shu2024dont};\cite{gupta2024selfassessment};\cite{pan2023llms};
    	 , fill=color2!10, text width=10cm, align=left, 
    	]
    	[\textbf{Text Analysis for Responses} \\ 
    	 \cite{karra2022estimating}; \cite{hilliard2024eliciting};\cite{song2024identifying}, fill=color2!10, text width=10cm, align=left
    	]
    ]
    [Exhibition \\ (Section \ref{Exhibition}), fill=color3!40, minimum width=2cm, rounded corners
    	[\textbf{Editing LLM's Personality} \\ 
    	 \cite{pan2023llms};\cite{karra2022estimating};\cite{liu2024dynamic}; \cite{cui2023machine}; \cite{mao2024editing}; \\ \cite{li2023does}
              , fill=color3!10, text width=11.2cm, align=left
    	]
    	[\textbf{Inducing LLM's Personality} \\                            \cite{pan2023llms};\cite{ramirez2023controlling};\cite{huang2023chatgptMBTI};\cite{choi2024picle};\cite{klinkert2024driving}; \\ \cite{he2024afspp};\cite{petrov2024limited};\cite{xu2024academically};\cite{tan2024phantom};\cite{noever2023ai};\\ \cite{jiang2024personallm};\cite{huang2023chatgptMBTI};\cite{lacava2024open};\cite{jiang2023evaluating};\cite{safdari2023personality}; \\ \cite{shen2024decisionmaking};\cite{kovac2023large};\cite{weng2024controllm};\cite{stockli2024personification}; \\ \cite{frisch2024llm}; \cite{gu2023effectiveness}, fill=color3!10, text width=11.2cm, align=left
    	]
    ]
    [Recognition \\ (Section \ref{Recognition}), fill=color4!40, minimum width=2cm, rounded corners
    	[\textbf{Personality Recognition by LLMs} \\ 
     \cite{peters2024large};\cite{ganesan2023systematic};\cite{amin2023will};\cite{ji2023chatgpt};\cite{derner2023chatgpt}; \\ \cite{peters2023large};\cite{10463124};\cite{rao-etal-2023-chatgpt};\cite{yang2023psycot}
    	 , fill=color4!10, text width=10.2cm, align=left
    	]
    	[\textbf{LLM-enhanced Personality Recognition} \\ 
     \cite{hu2024llmvssmall};\cite{wen2024affective};\cite{amin2023chatgpts};\cite{cao2024large}
    	 , fill=color4!10, text width=10.2cm, align=left
    	]
    ]
  ]
\end{forest}
\caption{Taxonomy of current studies on Personality in LLMs}
\label{fig:overview}
\end{figure*}
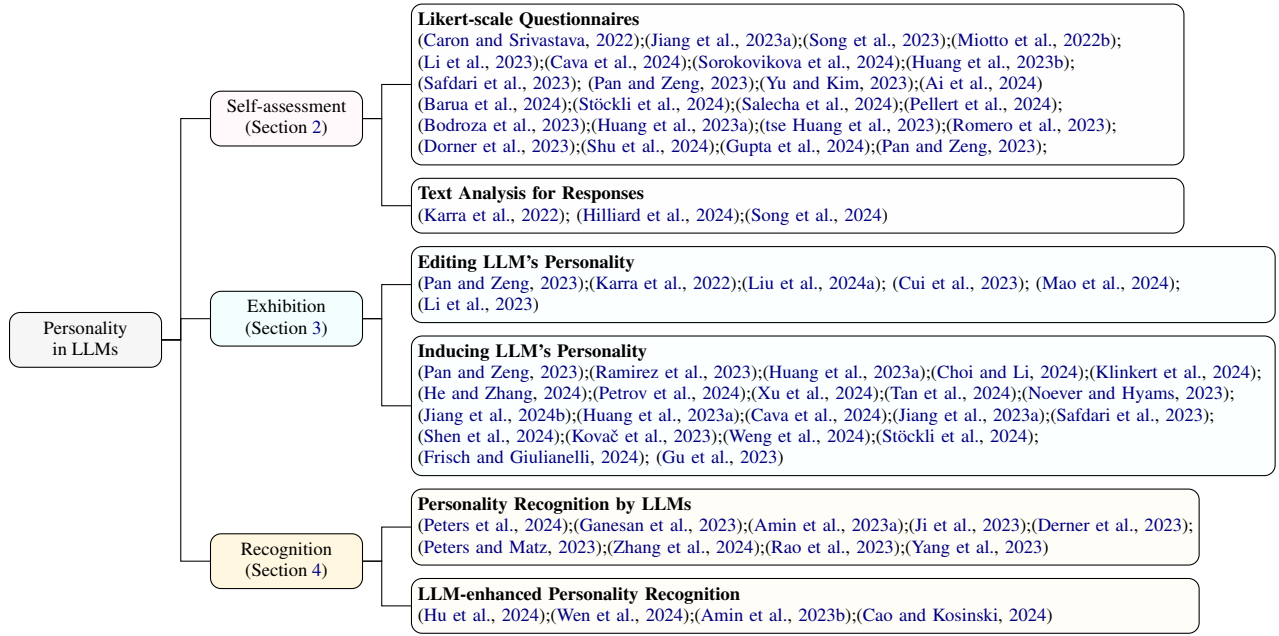

To fill in the research gap, we present a comprehensive review of up-to-date studies on personality in LLMs. We first propose a hierarchical taxonomy (at both the research problem level and the methodology level) to clearly organize the existing research, as shown in Figure \ref{fig:overview}. Specifically, we categorize personality in LLMs into three research problems based on the intrinsic characteristics and external manifestations: (1) Self-assessment, which measures the intrinsic personalities of LLMs, (2) Exhibition, which controls LLMs to exhibit specified personalities, and (3)  Recognition, which identifies personality traits from text content with LLMs. For each research problem, we further subdivide existing solutions based on their proposed methodologies.

In specific sections, we provide a thorough analysis of each problem with problem statements, motivations, and significance. Then, we conduct in-depth investigations and comparisons of the corresponding methods. Furthermore, we consolidate the findings and identify the open challenges revealed in current research. To facilitate researchers and developers, we also collect publicly available resources, including personality inventories, code repositories, and datasets. Lastly, we discuss potential future research directions and practical applications of personality in LLMs.

To summarize, the main contributions of our work are summarized as follows:

\begin{itemize}
\item \textbf{First Comprehensive Survey}: To the best of our knowledge, this is the first comprehensive survey of the latest studies on personality in LLMs. 

\item \textbf{Clear Hierarchical Taxonomy}: We propose a hierarchical taxonomy to clearly organize the literature at both the research problem level and the methodology level.
 
\item \textbf{Extensive Resource Collection}: We collect and summarize extensive publicly available resources to facilitate researchers and developers, including personality inventories, code repositories, and datasets, as shown in Appendix \ref{Open-sourced Resources}. 

\item \textbf{Promising Future Trends}: We summarize research findings and open challenges in current studies, and further discuss promising future research trends and potential application scenarios of personality in LLMs.

\end{itemize}




\section{LLM's Personality Self-assessment}
\label{Self-assessment}

\textit{Does a LLM possess a stable personality trait?} This question arises from LLMs' impressive human-like conversational experience, which drives researchers to investigate whether LLMs have acquired intrinsic personality traits in pre-training \cite{pellert2022ai}. Some studies \cite{huang2023chatgptMBTI, pan2023llms} suggest that ChatGPT approximates a consistent ENFJ personality type in MBTI, while others argue that LLMs' personalities are unstable \cite{pellert2022ai,miotto2022gpt}. The discrepancy may stem from variations in assessment methodology or the adaptability of LLMs to different contexts. Understanding LLMs' personalities not only helps foster engaging and empathetic interactions but also plays a crucial role in uncovering latent biases, mitigating the biases, and thereby enhancing the fairness and accuracy of LLMs \cite{karra2022estimating}. 

\subsection{Problem Statement}
LLM's Personality Self-assessment is stated as: \textbf{How to measure LLMs' intrinsic personality traits from their text responses?} Existing studies solving this problem are based on two important assumptions: (1) \textbf{Existence}: LLMs have acquired intrinsic personalities in pre-training, and (2) \textbf{Measurability}: psychometrics designed for human personality analysis are also applicable to LLMs.

When assessing human personality, researchers commonly employ Likert-scale personality questionnaires and written material analysis. Similarly, existing studies conduct two main approaches to investigate the intrinsic personality traits of LLMs: (1) prompting LLMs with Likert scale personality questionnaires, and (2) analyzing the text responses of LLMs under specific tasks. We will introduce these methods in the following content.

\subsection{Likert scale questionnaires}

Likert scale personality questionnaires (\textit{e.g.}, Table \ref{BFI_example}) consist of a series of multiple-choice questions (MCQs) that translate respondents' selections into numerical scores to assess personality traits, which are commonly used in the social sciences. Although MCQs are natural for humans, LLMs are designed for open-ended text input and output, making it difficult to conduct MCQs directly. Therefore, current researchers try various methods to prompt LLMs with questionnaires and extract the selected options from their text responses to derive the personality assessment results.

The most straightforward way is to prompt LLMs
 directly with questionnaire items and options \cite{song2023large,safdari2023personality,frisch2024llm} for personality test. 
However, it requires additional approaches, such as regular expressions analysis \cite{jiang2023evaluating}, designing parsers \cite{li2023does}, or analyzing the probability of tokens\cite{pan2023llms} to extract the answer options from the LLMs' text responses. Faced with this issue, some studies \cite{lacava2024open,stockli2024personification} found that adding instructive task descriptions and constraints in prompts, such as \textit{You will be provided a question ... to test your personality}, can facilitate obtaining the answer options.

Besides, since most LLMs, such as ChatGPT and Llama-chat, are configured to decline queries about personal opinions and experiences, this poses challenges for LLMs responding to personality questionnaires. To eliminate this constraint, researchers attempted to instruct LLMs to respond with only a number within the Likert-scale levels \cite{huang2023chatgpt}, rephrase the items into the third person plural \cite{miotto2022gpt3}, or add the role description \cite{sorokovikova2024llms}.

In addition to the mainstream generative LLMs, researchers also assessed the personalities of earlier pre-trained large language models by reformulating questionnaires into Natural Language Understanding (NLU) tasks. For instance, \citet{caron2022identifying} modifies the questionnaire by incorporating masked positions and prompts BERT to fill the answer options into them. \cite{pellert2022ai} employs the natural language inference (NLI) techniques to enable models such as DeBERTa to identify the most appropriate options corresponding to the items in the questionnaires.

\begin{table}[t]
    \centering
    \linespread{1.2}
    \renewcommand{\arraystretch}{1.2}
    \begin{tabular}{p{7cm}}
 	\toprule
 	\textit{I see myself as someone who is helpful and unselfish with others.}   \\
	\hline
	\textit{1 = Disagree strongly} \\
	\textit{2 = Disagree a little} \\
	\textit{3 = Neither agree nor disagree} \\
	\textit{4 = Agree a little} \\
	\textit{5 = Agree strongly} \\
	\hline
	\textit{Please write a number to indicate the extent to which you agree or disagree with that statement.}\\
	\bottomrule
    \end{tabular}
    \caption{The 7-th item in the BFI \cite{john1991big}.}
    \label{BFI_example}
\end{table}

\subsection{Text response analysis}

Despite most studies utilizing questionnaires to assess the personalities of LLMs, some researchers \cite{dorner2023personality, huang2023revisiting} still question whether LLMs, which are primarily designed for generating text content, can produce meaningful options in questionnaires. Therefore, reserchers conduct text analysis in semantic or linguistic perspectives on LLMs' responses to determine their personalities.

One direct method is to classify personality based on LLMs' responses. \cite{karra2022estimating} classifies the LLMs' responses to a personality questionnaire into the Big-five personality traits using a zero-shot classifier. Similarly, \cite{pellert2022ai} prompted the questions from personality inventories and conducted zero-shot classification on LLMs' responses to obtain their personality scores. Besides text responses to questionnaires,  answers to standard interview questions can also be analyzed to measure the Big Five personality traits of LLMs \cite{hilliard2024eliciting}.

Besides end-to-end text classifiers, Linguistic Inquiry and Word Count (LIWC, \citet{pennebaker2001linguistic}), a text analysis tool for personality analysis is also adopted for personality self-assessment of LLMs \cite{frisch2024llm,gu2023effectiveness,jiang2023personallm}. Vignette tests \cite{kwantes2016assessing} can also conducted by LLMs for personality assessment. In \cite{jiang2023evaluating}, LLMs are prompted with a description of a real-world scenario, followed by an open question and instructions for a short essay. Then, human participants were recruited to assess  LLMs' responses for personality reflection.

\subsection{Assessment Results Analysis}

Based on the various assessment methods introduced above, researchers have obtained various results on the personality of LLMs. These differences depend on a variety of factors: assessment approaches, prompting settings, model versions, hyperparameters, and so on. Nevertheless, several studies \cite{li2023does,huang2023chatgpt} agree on a tendency towards the Dark Triad traits in multiple LLMs, necessitating more rigorous research on the safety of these models.

The diversity of the assessment results also encourages researchers to investigate the robustness of the assessments. Although multiple terms are used and interpreted, such as reliability \cite{huang2023revisiting}, stability \cite{shu2024dont}, self-consistency \cite{pellert2022ai}, and validity \cite{romero2023gpt}, we summarize these terms into two perspectives: \textbf{Reliability}, which refers to the consistency and stability of assessment results over multiple repetitions; and \textbf{Validity}, which refers to the extent to which a test measures what it claims to measure, in other words, whether the personality assessment approaches used are indeed applicable to LLMs \cite{dorner2023personality}.  

\subsubsection{Reliability}
Several studies have reported a high reliability in LLMs' personality self-assessments. \cite{huang2023revisiting} conducted a comprehensive analysis across 2,500 experiments in different settings, demonstrating that GPT-3.5-turbo exhibits consistent behavior in responses to the Big Five Inventory. Similarly, \cite{huang2023chatgptMBTI} shows that ChatGPT consistently exhibits the ENFJ personality type across diverse languages, prompts, question orders, and rephrased inquiries in assessments.

However, not all findings are in agreement. \cite{li2023does} identified instances of conflicting answers and discrepancies in the responses generated by LLMs attributable to variations in the order of questionnaire options within prompts. Similarly, \cite{gupta2024selfassessment} identified inconsistent results of ChatGPT and Llama-2 across equivalent prompts in differing option presentations. Besides, \cite{song2023large} observed an inherent bias within LLMs, leading to a tendency to produce identical answers irrespective of the context.

Besides the differences in assessment approaches, LLMs' personalities are also observed to fluctuate with the temperature values \cite{miotto2022gpt3,huang2023chatgpt,barua2024psychology}. Larger parameter volumes and Supervised Fine-Tuning (SFT) can enhance the LLMs' assessment reliability \cite{serapiogarcía2023personality}.

\subsubsection{Validity}

Apart from reliability, there's also no consensus on the validity of personality assessment for LLMs.\cite{jiang2023evaluating} incorporated a validity test by prompting LLMs to explain the reason for selecting particular options in questionnaires. The results indicated that LLMs displayed a strong understanding of the questionnaire items, highlighting their assessment validity. Nonetheless, upon exhaustive analysis, \cite{dorner2023personality} found that the LLMs' personality assessment results did not exhibit the intended patterns similar to those observed in human answers. Therefore, the assumption of \textbf{Measurability} might be questionable.


\subsection{Findings and Open Challenges}

\noindent
\textbf{Finding 1: How do people assess LLMs' personalities?} Although some studies use text classification or linguistic tools to infer LLMs' personalities from text responses, most work still relies on prompt engineering for instructing LLMs to complete questionnaires for personality assessment.  

\noindent
\textbf{Finding 2: What are the assessment results?} Due to the diversity in assessment methods, even for the same LLM, there is no consensus on personality assessment results. Nonetheless, multiple studies agree that LLMs often exhibit darker traits than humans. 

\noindent
\textbf{Finding 3: Are the assessments meaningful?}
Although existing work conducts multiple repetitions of experiments in different settings to obtain more reliable results, the validity of measuring LLM's personality with psychometrics designed for humans has not yet been verified.

\noindent
\textbf{Challenge 1: Unified assessment approach} Due to the differences in different LLMs handling inputs and outputs, it's difficult to have a unified assessment approach (\textit{i.e.}, with the same inputs and answer extraction methods) that yields valid results across different LLMs. 

\noindent
\textbf{Challenge 2: Dark traits elimination in LLMs}: LLMs are uncovered to often exhibit darker or more negative traits compared to the human average. This may cause misinformation, ethical concerns, and potential harm to users' mental health. Effectively eliminating these traits while retaining the interactive capabilities of LLMs remains an open question.
\section{LLM's Personality Exhibition}
\label{Exhibition}
The capacity to exhibit diverse personalities is crucial for LLMs to satisfy users' needs in various application scenarios \cite{jiang2023evaluating}. Besides, enabling LLMs to adapt their personality traits to changing environmental factors contributes to dynamic AI systems that better align with users' changing needs and preferences \cite{karra2022estimating}. More importantly, pioneering studies \cite{gehman2020realtoxicityprompts,bender2021dangers,bommasani2021opportunities,tamkin2021understanding} observe that LLMs are prone to generate potentially harmful content due to unavoidable toxic data in pre-training. Adjusting the personality traits of LLMs effectively reduces the chance of toxic content by influencing the text's tone, style, and substance \cite{li2023does}. 

\subsection{Problem Statement}

LLM's personality exhibition can be stated as: \textbf{How to control LLMs to reflect the specified personality traits in the generated text content?} Current approaches to solving this problem are mainly categorized into:
\textbf{Editing}, which modifies the model parameters of LLMs to alter the potential intrinsic personality of LLMs acquired from pre-training; and \textbf{Inducing}, which fixes the LLM but utilizes prompt engineering to induce LLMs to exhibit specific personalities.

\subsection{Editing LLM's Personality}

One straightforward method for shaping personalities of LLMs is altering the model parameters through continual pre-training or fine-tuning on specific corpora. \cite{pan2023llms} conduct continual pre-training on LLMs, finding that the type of training corpus (\textit{e.g.}, wiki, question-answering, or examination materials) can affect the MBTI type exhibited by LLMs, especially in the dimensions of T/F and J/P. While \cite{karra2022estimating} shows that the personalities of LLMs (GPT-2) can be altered by fine-tuning on auxiliary classification or generation tasks. Similarly, \cite{liu2024dynamic} constructed a personality-dialogue dataset to fine-tune LLMs on generating dialogue content aligned with specified personality traits, assessed by GPT-4. 
Although these methods show effectiveness, it is also suggested that traits unintended to change may also be modified inadvertently, leading to undesired personality exhibition \cite{karra2022estimating}. 


Besides continual pre-training and traditional fine-tuning, instruction fine-tuning \cite{ouyang2022training}, originally designed to boost LLMs to follow human instructions to perform various tasks, also gains a lot of attention in editing LLM's personality. \cite{li2023does} conducted instruction fine-tuning to GPT-3 using items from the BFI and their corresponding answers in higher \textit{agreeableness} and lower \textit{neuroticism}, leading to a more positive and emotionally stable personality exhibition. Besides questionnaires, \cite{cui2023machine} construct the Behavior dataset by employing ChatGPT to classify question-answering (QA) pairs in the original Alpaca dataset \cite{alpaca} by MBTI dimensions. Similarly, \cite{mao2024editing} leveraged GPT-4 to generate QA pairs in specific scenarios facilitated by psychology domain knowledge. These dataset facilitate instruction fine-tuning LLMs exhibiting specific personalities.

It is noteworthy that most of the aforementioned fine-tuning methods employ parameter-efficient fine-tuning techniques, utilizing either LoRA \cite{mao2024editing,cui2023machine,liu2024dynamic} or adapter modules \cite{liu2024dynamic}. These methods allow for adjustments to a small subset of LLMs' parameters to attain the desired results.

\subsection{Inducing LLM's Personality}
While the editing methods have demonstrated partial effectiveness, an alternative method that is more widely applied is to employ prompting techniques to induce LLMs to exhibit specific personalities. Following existing research \cite{pan2023llms}, we categorize the inducing methods into Explicit Prompting, which utilizes the explicit description or definition of personality as the prompt; and Implicit Prompting, which leverages demonstrative examples of how the specified personality is implied in real scenarios as the prompts in an in-context learning manner. 

\noindent
\textbf{Explicit Prompting}: 
According to the lexical hypothesis of personality \cite{cutler2022deep}, personality is defined by the descriptive words of humans. Numerous researchers \cite{jiang2023evaluating, safdari2023personality, weng2024controllm, stockli2024personification} adopt descriptive adjectives of personality from psychological findings as the prompt content to elicit personality exhibition in LLMs. Although these adjectives can precisely describe specified personality traits, they struggle to provide detailed guidance on exhibiting specific personalities.

Concurrently, there are also studies prompting LLMs with descriptions of personalities \cite{pan2023llms, tan2024phantom, huang2023chatgptMBTI, lacava2024open, jiang2023personallm, kovac2023large} or interpretations of personality from psychological questionnaires \cite{noever2023ai}. Nevertheless, due to the sensitivity of LLMs to prompts, the efficacy of such methods is also affected by the content quality and the phrasings. 

\noindent
\textbf{Implicit Prompting}: Although explicit prompting provides clear personality descriptions, it lacks precise guidance on how personalities are manifested in specific scenarios. In implicit prompting, QA pairs in personality questionnaires encapsulate the manifestations or preferences of personality across a variety of potential scenarios, which are natural demonstrative examples for LLMs \cite{pan2023llms,huang2023chatgptMBTI,klinkert2024driving}.
Besides, demographics or profiles of typical individuals with specific personalities are also utilized to guide LLMs in exhibiting the inherent personalities \cite{huang2023chatgptMBTI,he2024afspp,petrov2024limited}.
Moreover, some studies use concrete examples, such as restaurant reviews \cite{ramirez2023controlling} or social network behaviors \cite{he2024afspp} in specific personalities, to induce LLMs exhibiting behaviors that align with the examples.

\subsection{Findings and Open Challenges}

\noindent
\textbf{Finding 1: Which method is more effective in LLM's personality exhibition?} Given the diverse methods and datasets for different LLMs, it is challenging to deduce a universal conclusion. However, faced with extensive parameters in LLMs, inducing with prompts appears to be a more practical way. Moreover, some studies (\textit{e.g.,} \citet{mao2024editing}) have demonstrated that inducing methods outperform the editing methods across most metrics on the same LLMs. 

\noindent
\textbf{Finding 2: What is the performance of LLMs' personality exhibition in current studies?} In current studies, precisely controlling an LLM to exhibit a composite personality is still a relatively challenging task \cite{lacava2024open, safdari2023personality}. However, researchers \cite{huang2023chatgptMBTI,pan2023llms,jiang2023personallm} show that modifying certain dimensions or facets of personality is more feasible. 

\noindent
\textbf{Challenge 1: Inconsistency} Despite existing studies have validated partial effectiveness of their personality exhibition approaches, some studies \cite{ai2024cognition,song2024identifying} indicate a misalignment between the exhibited personalities in evaluation and those in real-world scenarios. This highlights the need for context-aware evaluations and persistent controlling methods to ensure consistent personality exhibition. 

\noindent
\textbf{Challenge 2: Stability} Since inducing with prompts is proven effective in influencing LLM's personality exhibition, the personality exhibited by the LLM may also be affected by the context during interactions. Despite many LLMs using system-level prompts or safeguards to prevent user input effects, ensuring stable personality exhibitions during interactions remains an open challenge.


% ...

%
%
%

\section{Personality Recognition in LLM}
\label{Recognition}

Personality recognition is crucial and a longstanding research problem in both social science and computer science. Due to privacy concerns and the professional nature of personality analysis, obtaining sufficient annotated data for model training has always been a significant challenge \cite{wen2023desprompt}. LLMs' exceptional zero-shot ability have, to some extent, mitigated the issue of limited availability of labeled data. Besides, as LLMs can generate explanations to their output \cite{jiang2023evaluating}, the interpretability of the personality recognition results is also substantially enhanced. Consequently, researchers have become curious about Personality Recognition in LLM.

\subsection{Problem Statement}

Personality Recognition in LLMs is stated as \textbf{How to utilize LLMs recognize the personality traits from the given text content?} Current related research is primarily divided into two aspects: \textbf{Personality Recognition by LLMs}, which explores the zero-shot capabilities of LLMs for personality recognition; and  \textbf{LLM-enhanced Personality Recognition}, which utilizes LLMs to enhance other personality recognition models. 

\subsection{Personality Recognition by LLMs}

Inspired by the LLMs' zero-shot capabilities in NLP tasks, researchers directly input text content, such as social media posts \cite{ganesan2023systematic,peters2023large}, human written documents \cite{ji2023chatgpt,derner2023chatgpt}, or daily conversation \cite{peters2024large} as the prompts to LLMs for personality recognition. Their results show that though without additional training or fine-tuning, LLMs indeed perform well and can also provide natural language explanations of the results through text-based logical reasoning \cite{ji2023chatgpt}. Beyond traditional text-based personality recognition, researchers also use video transcripts as inputs for LLMs to enable personality recognition in more diverse contexts \cite{amin2023will,10463124}. Interestingly, it is also suggested that mimicking a user's acquaintance can enhance LLMs' personality recognition performance in the conversation scenarios \cite{peters2024large}. 

Some researchers also explored LLMs' comprehension of personality questionnaires for personality recognition. \cite{rao-etal-2023-chatgpt} investigate the ability of LLMs in personality recognition based on MBTI questionnaires by observing how LLMs correlate the answers with underlying personality traits. They showed that LLMs undergone Reinforcement Learning from Human Feedback (RLHF) have better performance in personality recognition. Besides, researchers \cite{yang2023psycot} proposed to prompt LLMs with the items from the personality questionnaire in a chain-of-thought manner, emulating the way individuals complete psychological questionnaires. Their method is validated to significantly improve the performance and robustness of GPT-3.5 in personality recognition.

\subsection{LLM-enhanced Personality Recognition}

Although most LLMs can easily outperform traditional NN models and pre-trained language models on personality recognition in the zero-shot setting, they still underperform state-of-the-art (SOTA) models that are specially trained for personality recognition \cite{ji2023chatgpt,ganesan2023systematic,amin2023will}. Therefore, researchers also attempt to use LLMs to enhance existing personality recognition models by augmenting the input data \cite{hu2024llmvssmall,wen2024affective,amin2023chatgpts} or providing additional features \cite{cao2024large}.

For example, when using traditional NN models to identify personality traits from social media posts, LLMs can generate additional analysis on these posts in the aspects of semantic, sentiment, and linguistic as augmentation \cite{hu2024llmvssmall}. Moreover, LLMs can also generate semantic interpretations of personality classification labels in the study above. In the context of personality recognition in conversations, LLMs can be engaged in affective analysis of the utterances to provide additional cues for personality recognition \cite{wen2024affective}. Besides, the personality analysis from ChatGPT about the given text can serve as features to assist machine learning models in personality recognition \cite{amin2023chatgpts}. Due to the rich information acquired during pre-training, even the word embeddings from GPT-3 for names of well-known figures can contribute to analyzing their personality traits \cite{cao2024large}.

\subsection{Findings and Open Challenges}

\noindent
\textbf{Finding 1: Can we directly apply LLMs for personality recognition?} LLMs exhibit superior performance but still underperform the SOTA models in personality recognition. So, in practical applications, LLMs without specialized training or fine-tuning are not suitable for directly obtaining results of personality recognition. 

\noindent
\textbf{Finding 2: How can LLMs facilitate existing personality recognition models?} LLMs can be utilized to provide additional information, such as explanations of the input, auxillary features, and description of labels, to facilitate traditional personality recognition models.

\noindent
\textbf{Challenge 1: Demographic Biases} Despite the impressive performance in personality recognition, LLMs are observed to exhibit potential biases towards certain demographic attributes (\textit{i.e.}, the recognition performance of certain genders and ages is better than others) \cite{ji2023chatgpt,peters2023large}. Investigating the causes of this bias and eliminating it to achieve more fair personality recognition results is an open challenge.

\noindent
\textbf{Challenge 2: Positivity Biases} As most LLMs are refined by RLHF to align with human preferences, they tend to assign socially desirable scores across key personality dimensions to the input. This propensity may render the results less authentic and convincing \cite{derner2023chatgpt}. Correcting these positivity biases for more accurate results is also an open challenge.

\section{Future Directions}
Beyond addressing the open challenges in each problem, we also discuss some other promising future directions to provide insights for researchers to further advance this field.

\noindent
\textbf{Psychometrics Tailored to LLM:} 
The use of psychometrics designed for humans on LLMs has been questioned in existing studies \cite{dorner2023personality,shu2024dont}. Personalities exhibited by LLMs are determined by the pre-training process and the mechanism of response generation. To better assess and control the personalities exhibited in LLMs, it is necessary to adapt traditional psychometrics taking into account the understanding of LLMs. 

\noindent
\textbf{Life-long Monitoring of Personality in LLM:} One significant motivation for investigating Personality in LLMs is to create LLM-based conversational agents (CAs) that have long-term engagement with us. So, it's crucial to have life-long monitoring to ensure the LLM-based CAs consistently maintain a personality that aligns with our expectations. This monitoring might include LLM's personality self-assessment based on the conversation history, as well as the regulation of LLM's personality exhibition according to user feedback in interactions.

\noindent
\textbf{Multi-modal Personality in LLM-based Digital Human:} Personality exhibition is not limited to text. As the evolving of LLM-based digital humans, enabling them to recognize user personality through multimodal interactions and exhibit the specified personality through facial expressions or gestures can greatly enhance user experience.

\section{Applications}
Besides academic research, personality in LLMs has a wide range of practical applications. Effective personality self-assessment methods can help verify whether the developed LLM-based agents accurately play their assigned roles \cite{wang2024incharacter,de2024use}. Enabling LLMs to exhibit specific personalities can simulate data generation by different annotators \cite{kaszyca2023possible} or assist in developing intelligent tutoring systems \cite{liu2024personalityaware}. Moreover, personality recognition based on LLMs can be beneficial in psychiatric clinics \cite{cheng2023now} and in credit services \cite{yu4671511gpt} for identifying user risks. 
In addition,  LLM's personality exhibition can be applied to other scenarios beyond text, such as having multiple agents play different personalities for efficient multi-agent collaboration \cite{sun2024llmbased} or designing LLM-based personae in HCI scenarios \cite{mirjana2024hci}. Lastly, the personality in LLM can also serve as a gateway to exploring other capabilities of LLMs, such as decision-making \cite{shen2024decisionmaking,sreedhar2024simulating}, negotiation skills \cite{noh2024llms}, and cultural perspectives \cite{kovac2023large}.

\section{Conclusion}
This paper comprehensively reviews the latest studies of personality in LLM by systematically examining the three core research problems of self-assessment, exhibition, and recognition. We present an exhaustive analysis of each problem. Subsequently, we carry out detailed analyses and comparisons of the relevant methods. Lastly, we also collect publicly available resources, discuss potential future research directions, and summarize practical applications of personality in LLMs. 

As the first comprehensive survey on personality in LLMs, we cover the latest literature and aim to provide a good reference resource on this topic for both researchers and engineers. Additionally, we hope this survey can enhance mutual understanding between social sciences and computer science, fostering more valuable interdisciplinary research.

\newpage
\section*{Limitations}
Personality in Large Language Models (LLMs) is an interdisciplinary research area situated between computer science and social science. However, most of the studies we reviewed are from the perspective of computer science, which has also led to our taxonomy being more based on a computer science viewpoint. In our survey, we highlighted that some of the reviewed methods do not have a solid grounding in the social sciences. We have tried to find work on Personality in LLMs within the social science domain but with limited success. At present, there appears to be a research gap in this area. We hope our survey can attract researchers from the social sciences to contribute more rational research methodologies from social science perspectives to Personality in LLMs.

\bibliography{anthology}

\begin{thebibliography}{110}
\expandafter\ifx\csname natexlab\endcsname\relax\def\natexlab#1{#1}\fi

\bibitem[{Ai et~al.(2024)Ai, He, Zhang, Zhu, Hao, Yu, Chen, and Wang}]{ai2024cognition}
Yiming Ai, Zhiwei He, Ziyin Zhang, Wenhong Zhu, Hongkun Hao, Kai Yu, Lingjun Chen, and Rui Wang. 2024.
\newblock \href {http://arxiv.org/abs/2402.14679} {Is cognition and action consistent or not: Investigating large language model's personality}.

\bibitem[{Amin et~al.(2023{\natexlab{a}})Amin, Cambria, and Schuller}]{amin2023will}
Mostafa~M Amin, Erik Cambria, and Bj{\"o}rn~W Schuller. 2023{\natexlab{a}}.
\newblock Will affective computing emerge from foundation models and general artificial intelligence? a first evaluation of chatgpt.
\newblock \emph{IEEE Intelligent Systems}, 38(2):15--23.

\bibitem[{Amin et~al.(2023{\natexlab{b}})Amin, Cambria, and Schuller}]{amin2023chatgpts}
Mostafa~M. Amin, Erik Cambria, and Björn~W. Schuller. 2023{\natexlab{b}}.
\newblock \href {http://arxiv.org/abs/2307.04648} {Can chatgpt's responses boost traditional natural language processing?}

\bibitem[{Ashton and Lee(2009)}]{ashton2009hexaco}
Michael~C Ashton and Kibeom Lee. 2009.
\newblock The hexaco--60: A short measure of the major dimensions of personality.
\newblock \emph{Journal of personality assessment}, 91(4):340--345.

\bibitem[{Barua et~al.(2024)Barua, Brase, Dong, Hitzler, and Vasserman}]{barua2024psychology}
Adrita Barua, Gary Brase, Ke~Dong, Pascal Hitzler, and Eugene Vasserman. 2024.
\newblock On the psychology of gpt-4: Moderately anxious, slightly masculine, honest, and humble.
\newblock \emph{arXiv preprint arXiv:2402.01777}.

\bibitem[{Bender et~al.(2021)Bender, Gebru, McMillan-Major, and Shmitchell}]{bender2021dangers}
Emily~M Bender, Timnit Gebru, Angelina McMillan-Major, and Shmargaret Shmitchell. 2021.
\newblock On the dangers of stochastic parrots: Can language models be too big?
\newblock In \emph{Proceedings of the 2021 ACM conference on fairness, accountability, and transparency}, pages 610--623.

\bibitem[{Bodroza et~al.(2023)Bodroza, Dinic, and Bojic}]{bodroza2023personality}
Bojana Bodroza, Bojana~M. Dinic, and Ljubisa Bojic. 2023.
\newblock \href {http://arxiv.org/abs/2306.04308} {Personality testing of gpt-3: Limited temporal reliability, but highlighted social desirability of gpt-3's personality instruments results}.

\bibitem[{Bommasani et~al.(2021)Bommasani, Hudson, Adeli, Altman, Arora, von Arx, Bernstein, Bohg, Bosselut, Brunskill et~al.}]{bommasani2021opportunities}
Rishi Bommasani, Drew~A Hudson, Ehsan Adeli, Russ Altman, Simran Arora, Sydney von Arx, Michael~S Bernstein, Jeannette Bohg, Antoine Bosselut, Emma Brunskill, et~al. 2021.
\newblock On the opportunities and risks of foundation models.
\newblock \emph{arXiv preprint arXiv:2108.07258}.

\bibitem[{Brown et~al.(2020)Brown, Mann, Ryder, Subbiah, Kaplan, Dhariwal, Neelakantan, Shyam, Sastry, Askell et~al.}]{brown2020language}
Tom Brown, Benjamin Mann, Nick Ryder, Melanie Subbiah, Jared~D Kaplan, Prafulla Dhariwal, Arvind Neelakantan, Pranav Shyam, Girish Sastry, Amanda Askell, et~al. 2020.
\newblock Language models are few-shot learners.
\newblock \emph{Advances in neural information processing systems}, 33:1877--1901.

\bibitem[{Cao and Kosinski(2024)}]{cao2024large}
Xubo Cao and Michal Kosinski. 2024.
\newblock Large language models know how the personality of public figures is perceived by the general public.
\newblock \emph{Scientific Reports}, 14(1):6735.

\bibitem[{Caron and Srivastava(2022)}]{caron2022identifying}
Graham Caron and Shashank Srivastava. 2022.
\newblock \href {http://arxiv.org/abs/2212.10276} {Identifying and manipulating the personality traits of language models}.

\bibitem[{Cava et~al.(2024)Cava, Costa, and Tagarelli}]{lacava2024open}
Lucio~La Cava, Davide Costa, and Andrea Tagarelli. 2024.
\newblock \href {http://arxiv.org/abs/2401.07115} {Open models, closed minds? on agents capabilities in mimicking human personalities through open large language models}.

\bibitem[{Chen et~al.(2024)Chen, Wang, Xu, Yuan, Zhang, Shi, Xie, Li, Yang, Zhu et~al.}]{chen2024persona}
Jiangjie Chen, Xintao Wang, Rui Xu, Siyu Yuan, Yikai Zhang, Wei Shi, Jian Xie, Shuang Li, Ruihan Yang, Tinghui Zhu, et~al. 2024.
\newblock From persona to personalization: A survey on role-playing language agents.
\newblock \emph{arXiv preprint arXiv:2404.18231}.

\bibitem[{Chen et~al.(2023)Chen, Wang, Jiang, Cai, Li, Chen, Wang, and Li}]{chen2023large}
Nuo Chen, Yan Wang, Haiyun Jiang, Deng Cai, Yuhan Li, Ziyang Chen, Longyue Wang, and Jia Li. 2023.
\newblock Large language models meet harry potter: A dataset for aligning dialogue agents with characters.
\newblock In \emph{Findings of the Association for Computational Linguistics: EMNLP 2023}, pages 8506--8520.

\bibitem[{Cheng et~al.(2023)Cheng, Chang, Chang, Wang, Liang, Kishimoto, Chang, Kuo, and Su}]{cheng2023now}
Szu-Wei Cheng, Chung-Wen Chang, Wan-Jung Chang, Hao-Wei Wang, Chih-Sung Liang, Taishiro Kishimoto, Jane Pei-Chen Chang, John~S Kuo, and Kuan-Pin Su. 2023.
\newblock The now and future of chatgpt and gpt in psychiatry.
\newblock \emph{Psychiatry and clinical neurosciences}, 77(11):592--596.

\bibitem[{Choi and Li(2024)}]{choi2024picle}
Hyeong~Kyu Choi and Yixuan Li. 2024.
\newblock \href {http://arxiv.org/abs/2405.02501} {Picle: Eliciting diverse behaviors from large language models with persona in-context learning}.

\bibitem[{Cui et~al.(2023)Cui, Lv, Wen, Tang, Tian, and Yuan}]{cui2023machine}
Jiaxi Cui, Liuzhenghao Lv, Jing Wen, Jing Tang, YongHong Tian, and Li~Yuan. 2023.
\newblock Machine mindset: An mbti exploration of large language models.
\newblock \emph{arXiv preprint arXiv:2312.12999}.

\bibitem[{Cutler and Condon(2022)}]{cutler2022deep}
Andrew Cutler and David~M Condon. 2022.
\newblock Deep lexical hypothesis: Identifying personality structure in natural language.
\newblock \emph{Journal of Personality and Social Psychology}.

\bibitem[{De~Raad(2000)}]{de2000big}
Boele De~Raad. 2000.
\newblock \emph{The big five personality factors: the psycholexical approach to personality.}
\newblock Hogrefe \& Huber Publishers.

\bibitem[{de~Winter et~al.(2024)de~Winter, Driessen, and Dodou}]{de2024use}
Joost~CF de~Winter, Tom Driessen, and Dimitra Dodou. 2024.
\newblock The use of chatgpt for personality research: Administering questionnaires using generated personas.
\newblock \emph{Personality and Individual Differences}, 228:112729.

\bibitem[{Derner et~al.(2023)Derner, Kučera, Oliver, and Zahálka}]{derner2023chatgpt}
Erik Derner, Dalibor Kučera, Nuria Oliver, and Jan Zahálka. 2023.
\newblock \href {http://arxiv.org/abs/2312.16070} {Can chatgpt read who you are?}

\bibitem[{Devlin et~al.(2018)Devlin, Chang, Lee, and Toutanova}]{devlin2018bert}
Jacob Devlin, Ming-Wei Chang, Kenton Lee, and Kristina Toutanova. 2018.
\newblock Bert: Pre-training of deep bidirectional transformers for language understanding.
\newblock \emph{arXiv preprint arXiv:1810.04805}.

\bibitem[{Digman(1990)}]{digman1990personality}
John~M Digman. 1990.
\newblock Personality structure: Emergence of the five-factor model.
\newblock \emph{Annual review of psychology}, 41(1):417--440.

\bibitem[{Dorner et~al.(2023)Dorner, Suhr, Samadi, and Kelava}]{dorner2023personality}
Florian~E. Dorner, Tom Suhr, Samira Samadi, and Augustin Kelava. 2023.
\newblock \href {http://arxiv.org/abs/2311.05297} {Do personality tests generalize to large language models?}

\bibitem[{Eysenck et~al.(1985)Eysenck, Eysenck, and Barrett}]{eysenck1985revised}
Sybil~BG Eysenck, Hans~J Eysenck, and Paul Barrett. 1985.
\newblock A revised version of the psychoticism scale.
\newblock \emph{Personality and individual differences}, 6(1):21--29.

\bibitem[{Fleeson and Jayawickreme(2015)}]{fleeson2015whole}
William Fleeson and Eranda Jayawickreme. 2015.
\newblock Whole trait theory.
\newblock \emph{Journal of research in personality}, 56:82--92.

\bibitem[{Frisch and Giulianelli(2024)}]{frisch2024llm}
Ivar Frisch and Mario Giulianelli. 2024.
\newblock \href {http://arxiv.org/abs/2402.02896} {Llm agents in interaction: Measuring personality consistency and linguistic alignment in interacting populations of large language models}.

\bibitem[{Ganesan et~al.(2023)Ganesan, Lal, Nilsson, and Schwartz}]{ganesan2023systematic}
Adithya~V Ganesan, Yash~Kumar Lal, August~Hakan Nilsson, and H.~Andrew Schwartz. 2023.
\newblock \href {http://arxiv.org/abs/2306.01183} {Systematic evaluation of gpt-3 for zero-shot personality estimation}.

\bibitem[{Gehman et~al.(2020)Gehman, Gururangan, Sap, Choi, and Smith}]{gehman2020realtoxicityprompts}
Samuel Gehman, Suchin Gururangan, Maarten Sap, Yejin Choi, and Noah~A Smith. 2020.
\newblock Realtoxicityprompts: Evaluating neural toxic degeneration in language models.
\newblock \emph{arXiv preprint arXiv:2009.11462}.

\bibitem[{Goldberg et~al.(1999)}]{goldberg1999broad}
Lewis~R Goldberg et~al. 1999.
\newblock A broad-bandwidth, public domain, personality inventory measuring the lower-level facets of several five-factor models.
\newblock \emph{Personality psychology in Europe}, 7(1):7--28.

\bibitem[{Gosling et~al.(2003)Gosling, Rentfrow, and Swann~Jr}]{gosling2003very}
Samuel~D Gosling, Peter~J Rentfrow, and William~B Swann~Jr. 2003.
\newblock A very brief measure of the big-five personality domains.
\newblock \emph{Journal of Research in personality}, 37(6):504--528.

\bibitem[{Gu et~al.(2023)Gu, Degachi, Gen{\c{c}}, Chandrasegaran, and Verma}]{gu2023effectiveness}
Heng Gu, Chadha Degachi, U{\u{g}}ur Gen{\c{c}}, Senthil Chandrasegaran, and Himanshu Verma. 2023.
\newblock On the effectiveness of creating conversational agent personalities through prompting.
\newblock \emph{arXiv preprint arXiv:2310.11182}.

\bibitem[{Gupta et~al.(2024)Gupta, Song, and Anumanchipalli}]{gupta2024selfassessment}
Akshat Gupta, Xiaoyang Song, and Gopala Anumanchipalli. 2024.
\newblock \href {http://arxiv.org/abs/2309.08163} {Self-assessment tests are unreliable measures of llm personality}.

\bibitem[{He et~al.(2023)He, Fu, Yu, Wang, Li, Zhao, Song, Qi, Luo, Zou, and Yang}]{he2023psychological}
Tianyu He, Guanghui Fu, Yijing Yu, Fan Wang, Jianqiang Li, Qing Zhao, Changwei Song, Hongzhi Qi, Dan Luo, Huijing Zou, and Bing~Xiang Yang. 2023.
\newblock \href {http://arxiv.org/abs/2312.04578} {Towards a psychological generalist ai: A survey of current applications of large language models and future prospects}.

\bibitem[{He and Zhang(2024)}]{he2024afspp}
Zihong He and Changwang Zhang. 2024.
\newblock \href {http://arxiv.org/abs/2401.02870} {Afspp: Agent framework for shaping preference and personality with large language models}.

\bibitem[{Hilliard et~al.(2024)Hilliard, Munoz, Wu, and Koshiyama}]{hilliard2024eliciting}
Airlie Hilliard, Cristian Munoz, Zekun Wu, and Adriano~Soares Koshiyama. 2024.
\newblock \href {http://arxiv.org/abs/2402.08341} {Eliciting personality traits in large language models}.

\bibitem[{Hodo(2006)}]{hodo2006kaplan}
David~W Hodo. 2006.
\newblock Kaplan and sadock's comprehensive textbook of psychiatry.
\newblock \emph{American Journal of Psychiatry}, 163(8):1458--1458.

\bibitem[{Hu et~al.(2024)Hu, He, Wang, Zhao, Shao, and Nie}]{hu2024llmvssmall}
Linmei Hu, Hongyu He, Duokang Wang, Ziwang Zhao, Yingxia Shao, and Liqiang Nie. 2024.
\newblock \href {http://arxiv.org/abs/2403.07581} {Llmvssmall model? large language model based text augmentation enhanced personality detection model}.

\bibitem[{Huang et~al.(2023{\natexlab{a}})Huang, Wang, Lam, Li, Jiao, and Lyu}]{huang2023chatgptMBTI}
Jen-tse Huang, Wenxuan Wang, Man~Ho Lam, Eric~John Li, Wenxiang Jiao, and Michael~R Lyu. 2023{\natexlab{a}}.
\newblock Chatgpt an enfj, bard an istj: Empirical study on personalities of large language models.
\newblock \emph{arXiv preprint arXiv:2305.19926}.

\bibitem[{Huang et~al.(2023{\natexlab{b}})Huang, Wang, Li, Lam, Ren, Yuan, Jiao, Tu, and Lyu}]{huang2023chatgpt}
Jen-tse Huang, Wenxuan Wang, Eric~John Li, Man~Ho Lam, Shujie Ren, Youliang Yuan, Wenxiang Jiao, Zhaopeng Tu, and Michael~R Lyu. 2023{\natexlab{b}}.
\newblock Who is chatgpt? benchmarking llms' psychological portrayal using psychobench.
\newblock \emph{arXiv preprint arXiv:2310.01386}.

\bibitem[{Ji et~al.(2023)Ji, Wu, Zheng, Hu, Chen, and He}]{ji2023chatgpt}
Yu~Ji, Wen Wu, Hong Zheng, Yi~Hu, Xi~Chen, and Liang He. 2023.
\newblock Is chatgpt a good personality recognizer? a preliminary study.
\newblock \emph{arXiv preprint arXiv:2307.03952}.

\bibitem[{Jiang et~al.(2024{\natexlab{a}})Jiang, Sablayrolles, Roux, Mensch, Savary, Bamford, Chaplot, Casas, Hanna, Bressand et~al.}]{jiang2024mixtral}
Albert~Q Jiang, Alexandre Sablayrolles, Antoine Roux, Arthur Mensch, Blanche Savary, Chris Bamford, Devendra~Singh Chaplot, Diego de~las Casas, Emma~Bou Hanna, Florian Bressand, et~al. 2024{\natexlab{a}}.
\newblock Mixtral of experts.
\newblock \emph{arXiv preprint arXiv:2401.04088}.

\bibitem[{Jiang et~al.(2023{\natexlab{a}})Jiang, Xu, Zhu, Han, Zhang, and Zhu}]{jiang2023evaluating}
Guangyuan Jiang, Manjie Xu, Song-Chun Zhu, Wenjuan Han, Chi Zhang, and Yixin Zhu. 2023{\natexlab{a}}.
\newblock Evaluating and inducing personality in pre-trained language models.
\newblock In \emph{NeurIPS}.

\bibitem[{Jiang et~al.(2024{\natexlab{b}})Jiang, Zhang, Cao, Breazeal, Roy, and Kabbara}]{jiang2024personallm}
Hang Jiang, Xiajie Zhang, Xubo Cao, Cynthia Breazeal, Deb Roy, and Jad Kabbara. 2024{\natexlab{b}}.
\newblock \href {http://arxiv.org/abs/2305.02547} {Personallm: Investigating the ability of large language models to express personality traits}.

\bibitem[{Jiang et~al.(2023{\natexlab{b}})Jiang, Zhang, Cao, Kabbara, and Roy}]{jiang2023personallm}
Hang Jiang, Xiajie Zhang, Xubo Cao, Jad Kabbara, and Deb Roy. 2023{\natexlab{b}}.
\newblock Personallm: Investigating the ability of gpt-3.5 to express personality traits and gender differences.
\newblock \emph{arXiv preprint arXiv:2305.02547}.

\bibitem[{John et~al.(1991)John, Donahue, and Kentle}]{john1991big}
Oliver~P John, Eileen~M Donahue, and Robert~L Kentle. 1991.
\newblock Big five inventory.
\newblock \emph{Journal of personality and social psychology}.

\bibitem[{Johnson(2014)}]{JOHNSON201478}
John~A. Johnson. 2014.
\newblock \href {https://doi.org/https://doi.org/10.1016/j.jrp.2014.05.003} {Measuring thirty facets of the five factor model with a 120-item public domain inventory: Development of the ipip-neo-120}.
\newblock \emph{Journal of Research in Personality}, 51:78--89.

\bibitem[{Jonason and Webster(2010)}]{jonason2010dirty}
Peter~K Jonason and Gregory~D Webster. 2010.
\newblock The dirty dozen: a concise measure of the dark triad.
\newblock \emph{Psychological assessment}, 22(2):420.

\bibitem[{Jones and Paulhus(2014)}]{jones2014introducing}
Daniel~N Jones and Delroy~L Paulhus. 2014.
\newblock Introducing the short dark triad (sd3) a brief measure of dark personality traits.
\newblock \emph{Assessment}, 21(1):28--41.

\bibitem[{Karra et~al.(2022)Karra, Nguyen, and Tulabandhula}]{karra2022estimating}
Saketh~Reddy Karra, Son~The Nguyen, and Theja Tulabandhula. 2022.
\newblock Estimating the personality of white-box language models.
\newblock \emph{arXiv preprint arXiv:2204.12000}.

\bibitem[{Kaszyca et~al.(2023)Kaszyca, Kazienko, Koco{\'n}, Cichecki, Kochanek, and Szyd{\l}o}]{kaszyca2023possible}
Oliwier Kaszyca, Przemys{\l}aw Kazienko, Jan Koco{\'n}, Igor Cichecki, Mateusz Kochanek, and Dominka Szyd{\l}o. 2023.
\newblock Is it possible for chatgpt to mimic human annotator?

\bibitem[{Klinkert et~al.(2024)Klinkert, Buongiorno, and Clark}]{klinkert2024driving}
Lawrence~J. Klinkert, Stephanie Buongiorno, and Corey Clark. 2024.
\newblock \href {http://arxiv.org/abs/2402.14879} {Driving generative agents with their personality}.

\bibitem[{Kovač et~al.(2023)Kovač, Sawayama, Portelas, Colas, Dominey, and Oudeyer}]{kovac2023large}
Grgur Kovač, Masataka Sawayama, Rémy Portelas, Cédric Colas, Peter~Ford Dominey, and Pierre-Yves Oudeyer. 2023.
\newblock \href {http://arxiv.org/abs/2307.07870} {Large language models as superpositions of cultural perspectives}.

\bibitem[{Kwantes et~al.(2016)Kwantes, Derbentseva, Lam, Vartanian, and Marmurek}]{kwantes2016assessing}
Peter~J Kwantes, Natalia Derbentseva, Quan Lam, Oshin Vartanian, and Harvey~HC Marmurek. 2016.
\newblock Assessing the big five personality traits with latent semantic analysis.
\newblock \emph{Personality and Individual Differences}, 102:229--233.

\bibitem[{Lang et~al.(2011)Lang, John, L{\"u}dtke, Schupp, and Wagner}]{lang2011short}
Frieder~R Lang, Dennis John, Oliver L{\"u}dtke, J{\"u}rgen Schupp, and Gert~G Wagner. 2011.
\newblock Short assessment of the big five: Robust across survey methods except telephone interviewing.
\newblock \emph{Behavior research methods}, 43:548--567.

\bibitem[{Lee and Ashton(2018)}]{lee2018psychometric}
Kibeom Lee and Michael~C Ashton. 2018.
\newblock Psychometric properties of the hexaco-100.
\newblock \emph{Assessment}, 25(5):543--556.

\bibitem[{Li et~al.(2023)Li, Li, Joty, Liu, Huang, Qiu, and Bing}]{li2023does}
Xingxuan Li, Yutong Li, Shafiq Joty, Linlin Liu, Fei Huang, Lin Qiu, and Lidong Bing. 2023.
\newblock \href {http://arxiv.org/abs/2212.10529} {Does gpt-3 demonstrate psychopathy? evaluating large language models from a psychological perspective}.

\bibitem[{Liu et~al.(2024{\natexlab{a}})Liu, Gu, Zheng, Xiang, Wu, Fu, and He}]{liu2024dynamic}
Jianzhi Liu, Hexiang Gu, Tianyu Zheng, Liuyu Xiang, Huijia Wu, Jie Fu, and Zhaofeng He. 2024{\natexlab{a}}.
\newblock \href {http://arxiv.org/abs/2404.07084} {Dynamic generation of personalities with large language models}.

\bibitem[{Liu et~al.(2024{\natexlab{b}})Liu, Yin, Lin, and Chen}]{liu2024personalityaware}
Zhengyuan Liu, Stella~Xin Yin, Geyu Lin, and Nancy~F. Chen. 2024{\natexlab{b}}.
\newblock \href {http://arxiv.org/abs/2404.06762} {Personality-aware student simulation for conversational intelligent tutoring systems}.

\bibitem[{Mao et~al.(2024)Mao, Wang, Wang, Jiang, Xie, Huang, and Zhang}]{mao2024editing}
Shengyu Mao, Xiaohan Wang, Mengru Wang, Yong Jiang, Pengjun Xie, Fei Huang, and Ningyu Zhang. 2024.
\newblock \href {http://arxiv.org/abs/2310.02168} {Editing personality for large language models}.

\bibitem[{Miotto et~al.(2022{\natexlab{a}})Miotto, Rossberg, and Kleinberg}]{miotto2022gpt}
Maril{\`u} Miotto, Nicola Rossberg, and Bennett Kleinberg. 2022{\natexlab{a}}.
\newblock Who is gpt-3? an exploration of personality, values and demographics.
\newblock \emph{arXiv preprint arXiv:2209.14338}.

\bibitem[{Miotto et~al.(2022{\natexlab{b}})Miotto, Rossberg, and Kleinberg}]{miotto2022gpt3}
Marilù Miotto, Nicola Rossberg, and Bennett Kleinberg. 2022{\natexlab{b}}.
\newblock \href {http://arxiv.org/abs/2209.14338} {Who is gpt-3? an exploration of personality, values and demographics}.

\bibitem[{Mischel et~al.(2007)Mischel, Shoda, and Ayduk}]{mischel2007introduction}
Walter Mischel, Yuichi Shoda, and Ozlem Ayduk. 2007.
\newblock \emph{Introduction to personality: Toward an integrative science of the person}.
\newblock John Wiley \& Sons.

\bibitem[{Myers(1962)}]{Myers_1962}
Isabel~Briggs Myers. 1962.
\newblock \href {https://doi.org/10.1037/14404-000} {\emph{The Myers-Briggs Type Indicator: Manual (1962).}}
\newblock The Myers-Briggs Type Indicator: Manual (1962). Consulting Psychologists Press, Palo Alto, CA, US.

\bibitem[{Noever and Hyams(2023)}]{noever2023ai}
David Noever and Sam Hyams. 2023.
\newblock \href {http://arxiv.org/abs/2308.07326} {Ai text-to-behavior: A study in steerability}.

\bibitem[{Noh and Chang(2024)}]{noh2024llms}
Sean Noh and Ho-Chun~Herbert Chang. 2024.
\newblock \href {http://arxiv.org/abs/2405.05248} {Llms with personalities in multi-issue negotiation games}.

\bibitem[{Ouyang et~al.(2022)Ouyang, Wu, Jiang, Almeida, Wainwright, Mishkin, Zhang, Agarwal, Slama, Ray et~al.}]{ouyang2022training}
Long Ouyang, Jeffrey Wu, Xu~Jiang, Diogo Almeida, Carroll Wainwright, Pamela Mishkin, Chong Zhang, Sandhini Agarwal, Katarina Slama, Alex Ray, et~al. 2022.
\newblock Training language models to follow instructions with human feedback.
\newblock \emph{Advances in neural information processing systems}, 35:27730--27744.

\bibitem[{Pan and Zeng(2023)}]{pan2023llms}
Keyu Pan and Yawen Zeng. 2023.
\newblock Do llms possess a personality? making the mbti test an amazing evaluation for large language models.
\newblock \emph{arXiv preprint arXiv:2307.16180}.

\bibitem[{Paulhus et~al.(2020)Paulhus, Buckels, Trapnell, and Jones}]{paulhus2020screening}
Delroy~L Paulhus, Erin~E Buckels, Paul~D Trapnell, and Daniel~N Jones. 2020.
\newblock Screening for dark personalities.
\newblock \emph{European Journal of Psychological Assessment}.

\bibitem[{Pellert et~al.(2022)Pellert, Lechner, Wagner, Rammstedt, and Strohmaier}]{pellert2022ai}
Max Pellert, Clemens Lechner, Claudia Wagner, Beatrice Rammstedt, and Markus Strohmaier. 2022.
\newblock Ai psychometrics: Assessing the psychological profiles of large language models through psychometric inventories.

\bibitem[{Pellert et~al.(2023)Pellert, Lechner, Wagner, Rammstedt, and Strohmaier}]{pellert2023ai}
Max Pellert, Clemens~M Lechner, Claudia Wagner, Beatrice Rammstedt, and Markus Strohmaier. 2023.
\newblock Ai psychometrics: Assessing the psychological profiles of large language models through psychometric inventories.
\newblock \emph{Perspectives on Psychological Science}, page 17456916231214460.

\bibitem[{Pellert et~al.(2024)Pellert, Lechner, Wagner, Rammstedt, and Strohmaier}]{pellert2024ai}
Maximilian Pellert, Clemens~M. Lechner, Carmen Wagner, Beatrice Rammstedt, and Markus Strohmaier. 2024.
\newblock \href {https://doi.org/10.1177/17456916231214460} {Ai psychometrics: Assessing the psychological profiles of large language models through psychometric inventories}.
\newblock \emph{Perspectives on Psychological Science}, 0(0).

\bibitem[{Pennebaker et~al.(2001)Pennebaker, Francis, and Booth}]{pennebaker2001linguistic}
James~W Pennebaker, Martha~E Francis, and Roger~J Booth. 2001.
\newblock Linguistic inquiry and word count: Liwc 2001.
\newblock \emph{Mahway: Lawrence Erlbaum Associates}, 71(2001):2001.

\bibitem[{Pennebaker and King(1999)}]{pennebaker1999linguistic}
James~W Pennebaker and Laura~A King. 1999.
\newblock Linguistic styles: language use as an individual difference.
\newblock \emph{Journal of personality and social psychology}, 77(6):1296.

\bibitem[{Peters et~al.(2024)Peters, Cerf, and Matz}]{peters2024large}
Heinrich Peters, Moran Cerf, and Sandra~C Matz. 2024.
\newblock Large language models can infer personality from free-form user interactions.
\newblock \emph{arXiv preprint arXiv:2405.13052}.

\bibitem[{Peters and Matz(2023)}]{peters2023large}
Heinrich Peters and Sandra Matz. 2023.
\newblock \href {http://arxiv.org/abs/2309.08631} {Large language models can infer psychological dispositions of social media users}.

\bibitem[{Petrov et~al.(2024)Petrov, Serapio-García, and Rentfrow}]{petrov2024limited}
Nikolay~B Petrov, Gregory Serapio-García, and Jason Rentfrow. 2024.
\newblock \href {http://arxiv.org/abs/2405.07248} {Limited ability of llms to simulate human psychological behaviours: a psychometric analysis}.

\bibitem[{Prpa et~al.(2024)Prpa, Troiano, Wood, and Coady}]{mirjana2024hci}
Mirjana Prpa, Giovanni~Maria Troiano, Matthew Wood, and Yvonne Coady. 2024.
\newblock \href {https://doi.org/10.1145/3613905.3636293} {Challenges and opportunities of llm-based synthetic personae and data in hci}.
\newblock In \emph{Extended Abstracts of the 2024 CHI Conference on Human Factors in Computing Systems}, CHI EA '24, New York, NY, USA. Association for Computing Machinery.

\bibitem[{Radford et~al.(2019)Radford, Wu, Child, Luan, Amodei, Sutskever et~al.}]{radford2019language}
Alec Radford, Jeffrey Wu, Rewon Child, David Luan, Dario Amodei, Ilya Sutskever, et~al. 2019.
\newblock Language models are unsupervised multitask learners.
\newblock \emph{OpenAI blog}, 1(8):9.

\bibitem[{Ramirez et~al.(2023)Ramirez, Alsalihy, Aggarwal, Li, Wu, and Walker}]{ramirez2023controlling}
Angela Ramirez, Mamon Alsalihy, Kartik Aggarwal, Cecilia Li, Liren Wu, and Marilyn Walker. 2023.
\newblock Controlling personality style in dialogue with zero-shot prompt-based learning.
\newblock \emph{arXiv preprint arXiv:2302.03848}.

\bibitem[{Rao et~al.(2023)Rao, Leung, and Miao}]{rao-etal-2023-chatgpt}
Haocong Rao, Cyril Leung, and Chunyan Miao. 2023.
\newblock \href {https://doi.org/10.18653/v1/2023.findings-emnlp.84} {Can {C}hat{GPT} assess human personalities? a general evaluation framework}.
\newblock In \emph{Findings of the Association for Computational Linguistics: EMNLP 2023}, pages 1184--1194, Singapore. Association for Computational Linguistics.

\bibitem[{Romero et~al.(2023)Romero, Fitz, and Nakatsuma}]{romero2023gpt}
Peter Romero, Stephen Fitz, and Teruo Nakatsuma. 2023.
\newblock Do gpt language models suffer from split personality disorder? the advent of substrate-free psychometrics.

\bibitem[{Safdari et~al.(2023)Safdari, Serapio-Garc{\'\i}a, Crepy, Fitz, Romero, Sun, Abdulhai, Faust, and Matari{\'c}}]{safdari2023personality}
Mustafa Safdari, Greg Serapio-Garc{\'\i}a, Cl{\'e}ment Crepy, Stephen Fitz, Peter Romero, Luning Sun, Marwa Abdulhai, Aleksandra Faust, and Maja Matari{\'c}. 2023.
\newblock Personality traits in large language models.
\newblock \emph{arXiv preprint arXiv:2307.00184}.

\bibitem[{Salecha et~al.(2024)Salecha, Ireland, Subrahmanya, Sedoc, Ungar, and Eichstaedt}]{salecha2024large}
Aadesh Salecha, Molly~E Ireland, Shashanka Subrahmanya, Jo{\~a}o Sedoc, Lyle~H Ungar, and Johannes~C Eichstaedt. 2024.
\newblock Large language models show human-like social desirability biases in survey responses.
\newblock \emph{arXiv preprint arXiv:2405.06058}.

\bibitem[{Scheier and Carver(1985)}]{scheier1985self}
Michael~F Scheier and Charles~S Carver. 1985.
\newblock The self-consciousness scale: A revised version for use with general populations 1.
\newblock \emph{Journal of Applied Social Psychology}, 15(8):687--699.

\bibitem[{Serapio-García et~al.(2023)Serapio-García, Safdari, Crepy, Sun, Fitz, Romero, Abdulhai, Faust, and Matarić}]{serapiogarcía2023personality}
Greg Serapio-García, Mustafa Safdari, Clément Crepy, Luning Sun, Stephen Fitz, Peter Romero, Marwa Abdulhai, Aleksandra Faust, and Maja Matarić. 2023.
\newblock \href {http://arxiv.org/abs/2307.00184} {Personality traits in large language models}.

\bibitem[{Shen et~al.(2024)Shen, Xie, Zhang, and Xu}]{shen2024decisionmaking}
Chenglei Shen, Guofu Xie, Xiao Zhang, and Jun Xu. 2024.
\newblock \href {http://arxiv.org/abs/2402.18807} {On the decision-making abilities in role-playing using large language models}.

\bibitem[{Shu et~al.(2024)Shu, Zhang, Choi, Dunagan, Logeswaran, Lee, Card, and Jurgens}]{shu2024dont}
Bangzhao Shu, Lechen Zhang, Minje Choi, Lavinia Dunagan, Lajanugen Logeswaran, Moontae Lee, Dallas Card, and David Jurgens. 2024.
\newblock \href {http://arxiv.org/abs/2311.09718} {You don't need a personality test to know these models are unreliable: Assessing the reliability of large language models on psychometric instruments}.

\bibitem[{Song et~al.(2024)Song, Adachi, Feng, Lin, Yu, Li, Gupta, Anumanchipalli, and Kaur}]{song2024identifying}
Xiaoyang Song, Yuta Adachi, Jessie Feng, Mouwei Lin, Linhao Yu, Frank Li, Akshat Gupta, Gopala Anumanchipalli, and Simerjot Kaur. 2024.
\newblock \href {http://arxiv.org/abs/2402.14805} {Identifying multiple personalities in large language models with external evaluation}.

\bibitem[{Song et~al.(2023)Song, Gupta, Mohebbizadeh, Hu, and Singh}]{song2023large}
Xiaoyang Song, Akshat Gupta, Kiyan Mohebbizadeh, Shujie Hu, and Anant Singh. 2023.
\newblock \href {http://arxiv.org/abs/2305.14693} {Have large language models developed a personality?: Applicability of self-assessment tests in measuring personality in llms}.

\bibitem[{Sorokovikova et~al.(2024)Sorokovikova, Fedorova, Rezagholi, and Yamshchikov}]{sorokovikova2024llms}
Aleksandra Sorokovikova, Natalia Fedorova, Sharwin Rezagholi, and Ivan~P. Yamshchikov. 2024.
\newblock \href {http://arxiv.org/abs/2402.01765} {Llms simulate big five personality traits: Further evidence}.

\bibitem[{Soto and John(2017)}]{soto2017next}
Christopher~J Soto and Oliver~P John. 2017.
\newblock The next big five inventory (bfi-2): Developing and assessing a hierarchical model with 15 facets to enhance bandwidth, fidelity, and predictive power.
\newblock \emph{Journal of personality and social psychology}, 113(1):117.

\bibitem[{Sreedhar and Chilton(2024)}]{sreedhar2024simulating}
Karthik Sreedhar and Lydia Chilton. 2024.
\newblock \href {http://arxiv.org/abs/2402.08189} {Simulating human strategic behavior: Comparing single and multi-agent llms}.

\bibitem[{St{\"o}ckli et~al.(2024)St{\"o}ckli, Joho, Lehner, and Hanne}]{stockli2024personification}
Leandro St{\"o}ckli, Luca Joho, Felix Lehner, and Thomas Hanne. 2024.
\newblock The personification of chatgpt (gpt-4)—understanding its personality and adaptability.
\newblock \emph{Information}, 15(6):300.

\bibitem[{Sun et~al.(2024)Sun, Huang, and Pompili}]{sun2024llmbased}
Chuanneng Sun, Songjun Huang, and Dario Pompili. 2024.
\newblock \href {http://arxiv.org/abs/2405.11106} {Llm-based multi-agent reinforcement learning: Current and future directions}.

\bibitem[{Tamkin et~al.(2021)Tamkin, Brundage, Clark, and Ganguli}]{tamkin2021understanding}
Alex Tamkin, Miles Brundage, Jack Clark, and Deep Ganguli. 2021.
\newblock Understanding the capabilities, limitations, and societal impact of large language models.
\newblock \emph{arXiv preprint arXiv:2102.02503}.

\bibitem[{Tan et~al.(2024)Tan, Yeo, Wu, Xu, Jain, Chadha, Jaidka, Liu, and Ng}]{tan2024phantom}
Fiona~Anting Tan, Gerard~Christopher Yeo, Fanyou Wu, Weijie Xu, Vinija Jain, Aman Chadha, Kokil Jaidka, Yang Liu, and See-Kiong Ng. 2024.
\newblock \href {http://arxiv.org/abs/2403.02246} {Phantom: Personality has an effect on theory-of-mind reasoning in large language models}.

\bibitem[{Taori et~al.(2023)Taori, Gulrajani, Zhang, Dubois, Li, Guestrin, Liang, and Hashimoto}]{alpaca}
Rohan Taori, Ishaan Gulrajani, Tianyi Zhang, Yann Dubois, Xuechen Li, Carlos Guestrin, Percy Liang, and Tatsunori~B. Hashimoto. 2023.
\newblock Stanford alpaca: An instruction-following llama model.
\newblock \url{https://github.com/tatsu-lab/stanford_alpaca}.

\bibitem[{Touvron et~al.(2023)Touvron, Lavril, Izacard, Martinet, Lachaux, Lacroix, Rozi{\`e}re, Goyal, Hambro, Azhar et~al.}]{touvron2023llama}
Hugo Touvron, Thibaut Lavril, Gautier Izacard, Xavier Martinet, Marie-Anne Lachaux, Timoth{\'e}e Lacroix, Baptiste Rozi{\`e}re, Naman Goyal, Eric Hambro, Faisal Azhar, et~al. 2023.
\newblock Llama: Open and efficient foundation language models.
\newblock \emph{arXiv preprint arXiv:2302.13971}.

\bibitem[{tse Huang et~al.(2023)tse Huang, Wang, Lam, Li, Jiao, and Lyu}]{huang2023revisiting}
Jen tse Huang, Wenxuan Wang, Man~Ho Lam, Eric~John Li, Wenxiang Jiao, and Michael~R. Lyu. 2023.
\newblock \href {http://arxiv.org/abs/2305.19926} {Revisiting the reliability of psychological scales on large language models}.

\bibitem[{Wang et~al.(2024)Wang, Xiao, tse Huang, Yuan, Xu, Guo, Tu, Fei, Leng, Wang, Chen, Li, and Xiao}]{wang2024incharacter}
Xintao Wang, Yunze Xiao, Jen tse Huang, Siyu Yuan, Rui Xu, Haoran Guo, Quan Tu, Yaying Fei, Ziang Leng, Wei Wang, Jiangjie Chen, Cheng Li, and Yanghua Xiao. 2024.
\newblock \href {http://arxiv.org/abs/2310.17976} {Incharacter: Evaluating personality fidelity in role-playing agents through psychological interviews}.

\bibitem[{Wen et~al.(2023)Wen, Cao, Yang, Wang, Yang, and Liu}]{wen2023desprompt}
Zhiyuan Wen, Jiannong Cao, Yu~Yang, Haoli Wang, Ruosong Yang, and Shuaiqi Liu. 2023.
\newblock \href {https://doi.org/https://doi.org/10.1016/j.ipm.2023.103422} {Desprompt: Personality-descriptive prompt tuning for few-shot personality recognition}.
\newblock \emph{Information Processing \& Management}, 60(5):103422.

\bibitem[{Wen et~al.(2024)Wen, Cao, Yang, Yang, and Liu}]{wen2024affective}
Zhiyuan Wen, Jiannong Cao, Yu~Yang, Ruosong Yang, and Shuaiqi Liu. 2024.
\newblock \href {https://doi.org/10.1109/PerCom59722.2024.10494487} {Affective- nli: Towards accurate and interpretable personality recognition in conversation}.
\newblock In \emph{2024 IEEE International Conference on Pervasive Computing and Communications (PerCom)}, pages 184--193.

\bibitem[{Weng et~al.(2024)Weng, He, Liu, Liu, and Zhao}]{weng2024controllm}
Yixuan Weng, Shizhu He, Kang Liu, Shengping Liu, and Jun Zhao. 2024.
\newblock \href {http://arxiv.org/abs/2402.10151} {Controllm: Crafting diverse personalities for language models}.

\bibitem[{Xu et~al.(2024)Xu, Lin, Han, Sun, and Sun}]{xu2024academically}
Ruoxi Xu, Hongyu Lin, Xianpei Han, Le~Sun, and Yingfei Sun. 2024.
\newblock Academically intelligent llms are not necessarily socially intelligent.
\newblock \emph{arXiv preprint arXiv:2403.06591}.

\bibitem[{Yang et~al.(2023)Yang, Shi, Wan, Quan, Wang, Wu, and Wu}]{yang2023psycot}
Tao Yang, Tianyuan Shi, Fanqi Wan, Xiaojun Quan, Qifan Wang, Bingzhe Wu, and Jiaxiang Wu. 2023.
\newblock \href {http://arxiv.org/abs/2310.20256} {Psycot: Psychological questionnaire as powerful chain-of-thought for personality detection}.

\bibitem[{Yu and Kim(2023)}]{yu2023personality}
Byunggu Yu and Junwhan Kim. 2023.
\newblock Personality of ai.
\newblock \emph{arXiv preprint arXiv:2312.02998}.

\bibitem[{Yu et~al.(2023)Yu, Bai, and Chen}]{yu4671511gpt}
Li~Yu, Xuefei Bai, and Zhiwei Chen. 2023.
\newblock Gpt-lgbm: A chatgpt-based integrated framework for credit scoring with textual and structured data.
\newblock \emph{Available at SSRN 4671511}.

\bibitem[{Zhang et~al.(2024)Zhang, Koutsoumpis, Oostrom, Holtrop, Ghassemi, and Vries}]{10463124}
Tianyi Zhang, Antonis Koutsoumpis, Janneke~K. Oostrom, Djurre Holtrop, Sina Ghassemi, and Reinout E.~de Vries. 2024.
\newblock \href {https://doi.org/10.1109/TAFFC.2024.3374875} {Can large language models assess personality from asynchronous video interviews? a comprehensive evaluation of validity, reliability, fairness, and rating patterns}.
\newblock \emph{IEEE Transactions on Affective Computing}, pages 1--16.

\bibitem[{Zheng et~al.(2023)Zheng, Liao, Deng, and Nie}]{zheng2023building}
Zhonghua Zheng, Lizi Liao, Yang Deng, and Liqiang Nie. 2023.
\newblock Building emotional support chatbots in the era of llms.
\newblock \emph{arXiv preprint arXiv:2308.11584}.

\end{thebibliography}
\bibliographystyle{acl_natbib}

\newpage
\appendix

\section{Overview}
\label{Overview}
We present a holistic overview of the latest studies based on statistical results on our taxonomy. We have collected a total of 72 released scientific papers on personality in LLMs since 2022, encompassing investigations, methodologies, and applications. We clarify that the papers we reviewed are about psychological personality in LLMs. Although we are aware there are also extensive studies focusing on LLM-based role-playing agents \cite{chen2024persona}, we exclude them from this survey.

\begin{figure}[t]
    \centering
    \includegraphics[scale=0.3]{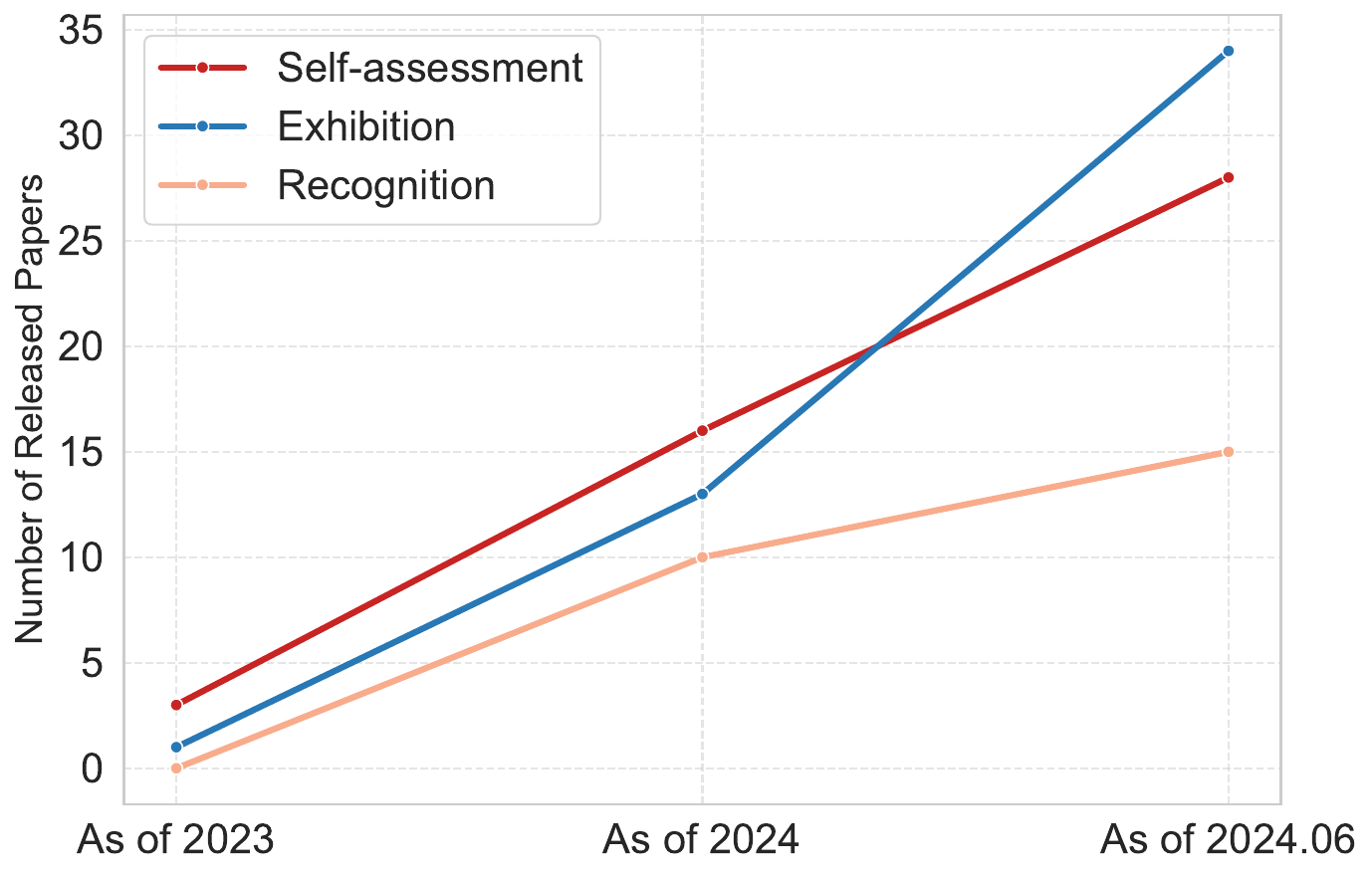}
    \caption{Trends of Personality in LLMs}
    \label{fig:trends}
\end{figure}

The number of papers in this emerging domain has been increasing annually, as shown in Figure \ref{fig:trends}. Even as of June 2024, the volume of publications has already surpassed that of the entire 2023. This indicates a growing interest in the field. Concurrently, we observe a substantial growth of work on LLMs' personality exhibition, underscoring the increasing focus on LLM-based interactions in various scenarios.  Following closely is research on LLMs' personality self-assessment, reflecting a sustained interest in exploring the intrinsic characteristics of LLMs. Compared to the two new research problems, there is a relatively less increase of personality recognition in LLM. This may be attributed to the fact that personality recognition, as a classical text classification problem, has been already widely studied with traditional methods. Nevertheless, personality recognition based on LLMs remains crucial in LLM-based interactions. The number of studies on this topic also continues to grow annually.

\subsection{Personality Models}
In reviewed literature, researchers commonly adopt personality models based on the trait theory \cite{fleeson2015whole}, where the personalities of individuals are defined as several aspects of stable and consistent patterns of behavior, emotion, and cognition. 

\begin{figure}[t]
    \centering
    \includegraphics[scale=0.3]{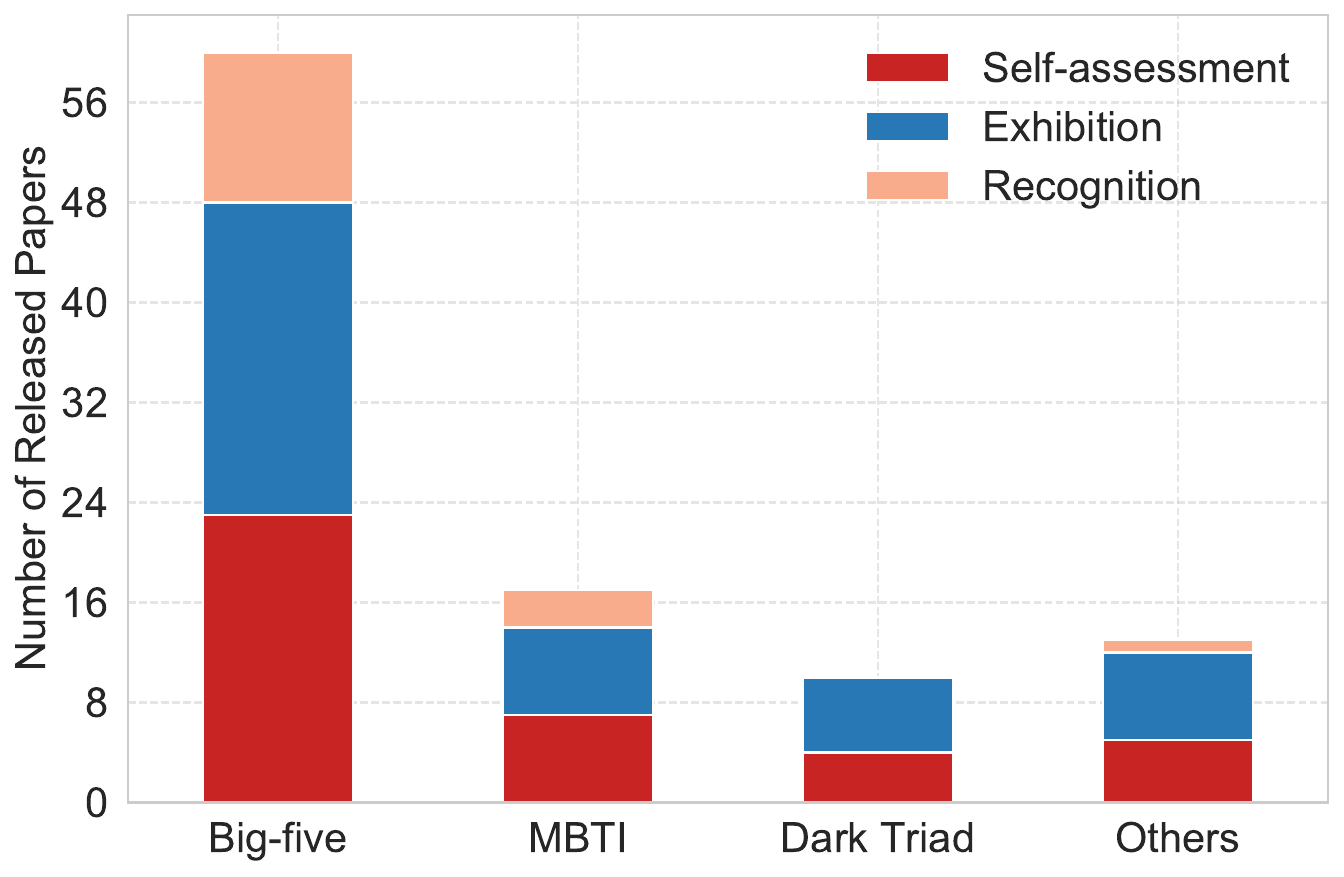}
    \caption{Personality Models in existing studies}
    \label{fig:personality models}
\end{figure}

As shown in Figure \ref{fig:personality models}, the most commonly adopted personality model among the existing three research problems is the Big-five model \cite{de2000big}, which includes five core dimensions: openness, conscientiousness, extraversion, agreeableness, and neuroticism. These dimensions capture various aspects of an individual's personality, ranging from their inclination towards new experiences to their level of emotional stability. Another widely recognized personality assessment model is the Myers-Briggs Type Indicator (\textbf{MBTI}) \cite{Myers_1962}, which categorizes individuals into one of 16 personality types based on their preferences in four dichotomous dimensions: extraversion/introversion, sensing/intuition, thinking/feeling, and judging/perceiving.

Besides the comprehensive personality models, researchers are also interested in the potential dark personality traits of LLMs, such as the Short Dark Triad-3 (SD-3, \citet{jones2014introducing}), or Dark Triad Dirty Dozen (DTDD, \citet{jonason2010dirty})  which measures Machiavellianism (a manipulative attitude), narcissism (excessive self-love), and psychopathy (lack of empathy), capturing the darker aspects of human nature.

In addition to the aforementioned personality models, researchers are also interested in (1) their variant models, such as the Eysenck Personality Questionnaire-Revised (EPQ-R) \cite{eysenck1985revised} assessing the dimensions of extraversion, neuroticism, and psychoticism, or the HEXACO model \cite{ashton2009hexaco} measuring honesty-humility in addition to the Big-five traits;  or (2) other psychological aspects in LLMs, such as motivations or interpersonal relationships \cite{huang2023chatgpt,bodroza2023personality}.

 \begin{figure*}[t]
     \centering
     \includegraphics[scale=0.4]{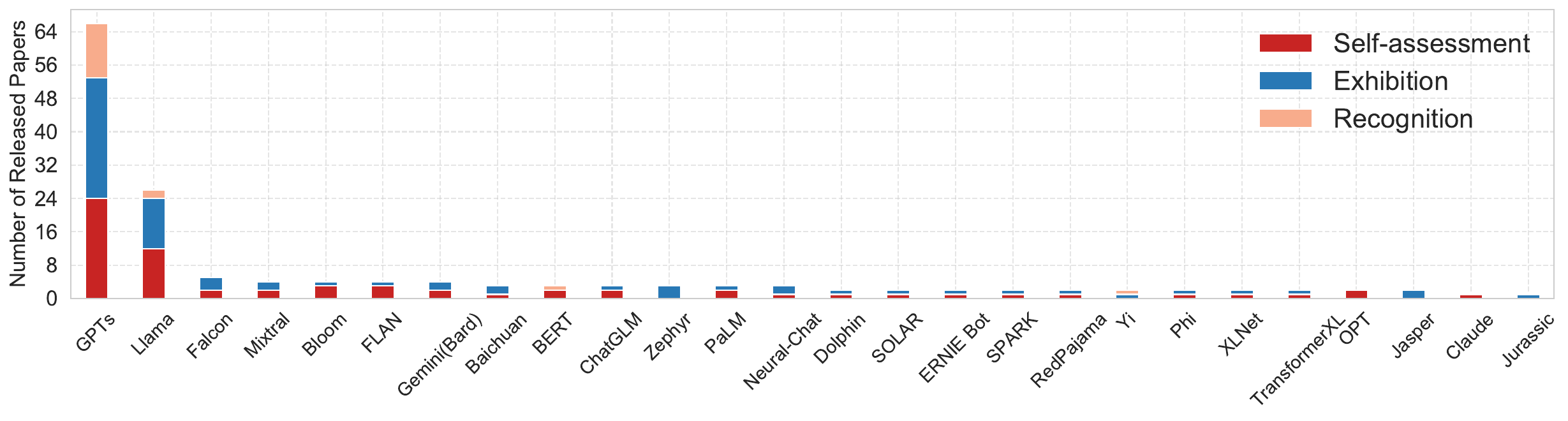}
     \caption{Investigated LLMs in existing studies}
     \label{fig:llms}
 \end{figure*}

\subsection{Large Language Models}

In existing studies, there is a significant interest in the performance of various LLMs on the three research problems, as shown in Figure \ref{fig:llms}. Among all the LLMs, the GPT series models have garnered the most attention from researchers. Though most of them are not directly open-sourced, extensive researchers (\textit{e.g.}, \cite{xu2024academically,karra2022estimating,jiang2023evaluating}) have explored their performance in personality understanding through API. Additionally, several well-known open-source LLMs, such as the Llama series \cite{touvron2023llama}, Mixtral \cite{jiang2024mixtral}, and Falcon\footnote{https://huggingface.co/docs/transformers/model\_doc/falcon}, have also attracted a lot of attention.

Among all LLMs, the task that researchers have focused on the most is personality self-assessment. Despite LLMs typically referring to models after the occurrence of GPT-3 \cite{brown2020language}, we also found that some pioneering work has already attempted to measure potential personality traits exhibited by BERT \cite{devlin2018bert} and GPT-2 \cite{radford2019language}.

\section{Open-sourced Resources}
\label{Open-sourced Resources}
\subsection{Personality Inventories}

\begin{table*}[h]
\centering
\small
\linespread{1}
\renewcommand{\arraystretch}{1.2}

\begin{tabular}{>{\raggedright\arraybackslash}m{5cm}m{4cm}cm{4cm}}

\toprule

 	\textbf{Personality Inventories} & \textbf{Measured Facets} & \textbf{\# Questions} & \textbf{Adopted in}\\
	\hline

Big Five Inventory \break  (BFI) \cite{john1991big} & Big-five & 44 & \cite{huang2023revisiting}; \cite{safdari2023personality}; \cite{ai2024cognition};\cite{pellert2022ai}; \cite{huang2023chatgpt};
\cite{li2023does};   \\

BFI-2 \break \cite{soto2017next}  & Big-five & 60 & \cite{li2023does}; \cite{huang2023chatgpt}; \cite{dorner2023personality}; \cite{safdari2023personality};\\

BFI-S \cite{lang2011short} & Big-five & 15 & \cite{jiang2023evaluating};\\


IPIP-NEO-120 \break \cite{JOHNSON201478} & Big-five (30 facets) & 120 & \cite{jiang2023evaluating} \\

IPIP-NEO\break  \cite{goldberg1999broad} & Big-five & 300 & \cite{safdari2023personality};\\
 
IPIP-50 (\href{https://ipip.ori.org/new\_ipip-50-item-scale.htm}{Link})& Big-five & 50 & \cite{dorner2023personality};\\

Ten Item Personality Measure \break (TIPI)  \cite{gosling2003very} & Big-five & 10 & \cite{romero2023gpt} \\

HEXACO \break \cite{ashton2009hexaco} & Honesty-humility and Big-five & 60 &  \cite{miotto2022gpt} \\

HEXACO-100 \break  \cite{lee2018psychometric} & Honesty-humility and Big-five & 100 &  \cite{bodroza2023personality}; \\

16Personalities (\href{https://www.16personalities.com/}{Link}) & MBTI & 60 & \cite{huang2023chatgptMBTI}; \cite{ai2024cognition}; \cite{rao-etal-2023-chatgpt}\\

Eysenck Personality Questionnaire-Revised \break (EPQ-R)  \cite{eysenck1985revised} & Extraversion, Neuroticism, Psychoticism, and Lying & 100 & \cite{huang2023chatgpt} \\

Dark Triad Dirty Dozen \break \cite{jonason2010dirty} 
	 & Machiavellianism , Narcissism, and Psychopathy (Dark Triad) & 12 & \cite{li2023does}, \cite{huang2023chatgpt}; \\

Short Dark Triad \break \cite{jones2014introducing} & Dark Triad  & 27 & \cite{bodroza2023personality} \\

Short Dark Tetrad \break (SD4)  \cite{paulhus2020screening} & Dark Triad and Sadism
& 28 & \cite{pellert2022ai} \\
Self-Consciousness Scales–Revised \break (SCS-R) \cite{scheier1985self} & Private self-consciousness, Public self-consciousness, and Social anxiety & 22 & \cite{bodroza2023personality} \\


\bottomrule
\end{tabular}
\caption{Personality inventories adopted by existing studies}
\label{tab: personality inventories}
\end{table*}

We have collected the personality inventories adopted in the papers we've reviewed (\textit{i.e.}, Likert-scale questionnaires) and listed them in Table \ref{tab: personality inventories}. We can see that the most commonly used inventory is the BFI. Similar to Figure \ref{fig:personality models}, most works focus on the Big-five personality model and its variants, such as HEXACO. Although many studies also focus on MBTI, we have not found many scales proposed in academic papers about MBTI. Therefore, current studies generally use questionnaires from 16Personalities\footnote{https://www.16personalities.com/}, a popular personality questionnaire website. Besides, many works also investigate the dark personality of LLMs.

It is evident that even when studying the same personality model, \textit{e.g}, Big-five, Dark Triad, different works will use a variety of personality inventories, and even some scales measure variants of these personality models. This reflects the diversity of psychometrics we mentioned in Introduction.

\subsection{Code Repositories of LLM's Personality Self-assessment}

Table \ref{tab:self-assessment} shows the publicly available code repositories in existing LLM's Personality Self-assessment works. Since the majority of studies focus on prompting engineering, which don't need much custom data or code, few have made their works publicly accessible. Among studies in Table \ref{tab:self-assessment}, apart from \cite{pellert2023ai} using NLI to assess the personalities of LLMs based on questionnaire questions and answers, all other works prompt LLMs with questionnaire items for personality assessment. We can also see that researchers exhibit interests in a wide range of LLMs, yet the majority still focus on LLMs in the GPT series.

\begin{table*}[h]
\centering
\small
\renewcommand{\arraystretch}{1.2}
\begin{center}

\begin{tabular}{>{\raggedright\arraybackslash}p{3cm}>{\raggedright\arraybackslash}p{4cm}>{\raggedright\arraybackslash}p{4cm}>{\raggedright\arraybackslash}p{4cm}}
\toprule
\textbf{Methods} & \textbf{Personality Models} & \textbf{Assessed LLMs} & \textbf{Download Links} \\
\hline

PsychoBench \break \cite{huang2023chatgpt} &  Big-five, EPQ-R, Short Dark Triad & GPT-3 (text-davinci-003); GPT-3.5-turbo,4; Llama-2-7B,13B & \href{https://github.com/CUHK-ARISE/PsychoBench}{\scriptsize{https://github.com/CUHK-ARISE/PsychoBench}} \\

\cite{shu2024dont} & MODEL-PERSONAS \break 39 instruments in 115 axis & GPT-2,3.5,4; Falcon-7B; BLOOMZ (all series); Llama2-7B,7B-chat,13B,13B-chat; RedPajama-7B; and FLAN-T5 (all series) & \href{https://github.com/orange0629/llm-personas}{\scriptsize{https://github.com/orange0629/llm-personas}}\\

\cite{miotto2022gpt} & HEXACO & GPT-3 & \href{https://github.com/ben-aaron188/who_is_gpt3}{\scriptsize{https://github.com/ben-aaron188/who\_is\_gpt3}} \\

\cite{romero2023gpt} & Big-five & GPT-3 & \href{https://osf.io/bf5c4/?view_only=f870d9e7258a4c2bb4430952f985b196}{\scriptsize{https://osf.io/bf5c4/}} \\

\cite{pellert2023ai} & Big-five & multilingualDeBERTa; DistilRoBERTa; BART; XLMRoBERTa; DeBERTa; DistilBART & \href{https://github.com/maxpel/psyai_materials}{\scriptsize{https://github.com/maxpel/psyai\_materials}} \\


\cite{stockli2024personification} & Big-five, MBTI & GPT4 &\href{https://github.com/AdritaBarua/2024-Psychology-of-GPT-4}{\scriptsize{https://github.com/AdritaBarua/2024-Psychology-of-GPT-4}}\\





\cite{jiang2023evaluating} & Big-five & GPT-3.5 & \href{https://github.com/jianggy/MPI}{\scriptsize{https://github.com/jianggy/MPI}} \\

\cite{safdari2023personality} & Big-five &  PaLM-62B; Flan-PaLM-8B,62B,540B; Flan-PaLMChilla-62B  & \href{https://github.com/google-research/google-research/tree/master/psyborgs}{\scriptsize{https://github.com/google-research/google-research/tree/master/psyborgs}}\\



\cite{huang2023revisiting} & Big-five & GPT-3.5-turbo & \href{https://github.com/CUHK-ARISE/LLMPersonality}{\scriptsize{https://github.com/CUHK-ARISE/LLMPersonality}} \\

\cite{pan2023llms} & MBTI & ChatGPT; GPT-4; Bloom-7B; Baichuan- 7B,13B; OpenLlama-7B-v2 & \href{https://github.com/HarderThenHarder/transformers_tasks/tree/main/LLM/llms_mbti}{\scriptsize{https://github.com/HarderThenHarder/ transformers\_tasks/tree/main/LLM/llms\_mbti}} \\


\cite{bodroza2023personality} & Big-five, HEXACO, SCS-R & GPT-3 (text-davinci-003) & \href{https://osf.io/2k458/?view\_only=6886694c6f8449488cfbc4e8f78ea2b0}{\scriptsize{https://osf.io/2k458}} \\

\bottomrule
\end{tabular}
\end{center}
\caption{Open source code repositories for LLM's personality self-assessment}
\label{tab:self-assessment}
\end{table*}

\subsection{Code Repositories of LLM's Personality Exhibition}

Table \ref{tab:exhibition_resources} shows the publicly available code repositories in existing LLM's Personality Exhibition works. Similar to LLM's personality self-assessment, most works in LLM's Personality Exhibition are prompting engineering in techniques, few have made their codes and data publicly accessible.
In Table \ref{tab:exhibition_resources}, while the majority of the work involves inducing the personality of LLMs with prompts, the evaluation to the induced results vary widely, encompassing Social Intelligence psychometric \cite{xu2024academically},  Theory-of-Mind reasoning tasks \cite{tan2024phantom}, response content analysis \cite{mao2024editing,jiang2023personallm,frisch2024llm,huang2023revisiting,wang2024incharacter}, and questionnaires \cite{jiang2023personallm,cui2023machine,huang2023revisiting,klinkert2024driving}.

\begin{table*}[h]
\centering
\small
\renewcommand{\arraystretch}{1.2}
\begin{center}

\begin{tabular}{>{\raggedright\arraybackslash}p{3cm}>{\raggedright\arraybackslash}p{4cm}>{\raggedright\arraybackslash}p{4cm}>{\raggedright\arraybackslash}p{4cm}}

\toprule
\textbf{Dataset} & \textbf{Personality Model} & \textbf{Descriptions} & \textbf{Download Links} \\
\hline
Situational Evaluation \break of Social Intelligence \break SESI \cite{xu2024academically} & Big-five & Prompt LLMs with personality descriptions with extents, evaluated by the proposed SESI & \href{https://github.com/RossiXu/social_intelligence_of_llms/}{\scriptsize{https://github.com/RossiXu/social\_in\break telligence\_of\_llms}} \\

PHAnToM \break \cite{tan2024phantom}& Big-five, Dark Triad & Prompt LLMs with personality descriptions, evaluated by Theory-of-Mind reasoning tasks & \href{https://anonymous.4open.science/r/PHAnToM/}{\scriptsize{https://anonymous.4open.science/r/P\break HAnToM/}} \\

PersonalityEdit \break \cite{mao2024editing}& Big-five (A, E, and N) & Utilize various methods in QA-based SFT and prompting for LLMs, evaluated by response content analysis & \href{https://github.com/zjunlp/EasyEdit/blob/main/examples/PersonalityEdit.md}{\scriptsize{https://github.com/zjunlp/EasyEdit/blob/mai\break n/examples/PersonalityEdit.md}} \\

PersonaLLM  \break \cite{jiang2023personallm}& Big-five & Prompt LLMs with personality descriptions, evaluated by questionnaires and story generation analysis & \href{https://github.com/hjian42/PersonaLLM}{\scriptsize{https://github.com/hjian42/PersonaLLM}} \\

MachineMindset \break  \cite{cui2023machine}& MBTI & Conduct two-phase SFT with QA datasets and DPO to LLMs, evaluated by questionnaires & \href{https://github.com/PKU-YuanGroup/Machine-Mindset}{\scriptsize{https://github.com/PKU-YuanGroup/Machine-Mindset}} \\


\cite{wang2024incharacter} & Big-five, MBTI &  Prompt personality descriptions and profiles to role-playing agents, assessed by interview analysis & \href{https://github.com/Neph0s/InCharacter/}{\scriptsize{https://github.com/Neph0s/InCharacter/}} \\

\cite{huang2023revisiting}& Big-five & Assign personalities via QA pairs, biography, and CoT-based portrayals, assessed with questionnaires in multiple formats and story generation analysis.  & \href{https://github.com/CUHK-ARISE/LLMPersonality}{\scriptsize{https://github.com/CUHK-ARISE/LLMPersonality}} \\

\cite{klinkert2024driving}& Big-five & Prompt LLMs with personality descriptions and numeric extents, assessed by a personality questionnaire & \href{https://gitlab.com/humin-game-lab/artificial-psychosocial-framework/-/tree/master/LLM\_Personality}{\scriptsize{https://gitlab.com/humin-game-lab/artificial-psychosocial-framework/-/tree/master/LLM\_Personality}}\\

\cite{frisch2024llm} & Big-five & Instruct LLMs with creative and analytical personalities to generate stories, evaluated by LIWC. & \href{https://github.com/ivarfresh/Interaction_LLMs}{\scriptsize{https://github.com/ivarfresh/\break Interaction\_LLMs}} \\

\bottomrule
\end{tabular}
\end{center}
\caption{Open source code repositories for LLM's personality exhibition}
\label{tab:exhibition_resources}
\end{table*}

\subsection{Datasets for Personality Recognition in LLM}

Table \ref{tab:recognition_resources} contains the open-source datasets adopted in existing studies on Personality Recognition in LLMs. Some datasets, such as Essays \cite{pennebaker1999linguistic}, PAN, and FriendsPersona, are classic text-based Personality Recognition datasets that have been explored by many studies before the emergence of LLMs. However, there are also some new datasets \cite{peters2024large} that were constructed facilitated by LLMs.

\begin{table*}[h]
\centering
\small
\renewcommand{\arraystretch}{1.2}
\begin{center}
\begin{tabular}{>{\raggedright\arraybackslash}p{3cm}>{\raggedright\arraybackslash}p{4cm}>{\raggedright\arraybackslash}p{4cm}>{\raggedright\arraybackslash}p{4cm}}
\toprule

\textbf{Datasets} & \textbf{Personality Models} & \textbf{Descriptions} & \textbf{Download Links} \\
\hline
\cite{rao-etal-2023-chatgpt}   & MBTI &  Questionnaires with 60 questions  &  \href{https://github.com/Kali-Hac/ChatGPT-MBTI}{\scriptsize{https://github.com/Kali-Hac/ChatGPT-MBTI}}\\

Essays \break \cite{pennebaker1999linguistic} &  Big-five & 2,468 self-report essays from more than 1,200 students & \href{https://github.com/preke/DesPrompt/tree/main/data/Essay}{\scriptsize{https://github.com/preke/DesPrompt/t\break ree/main/data/Essay}} \break *non-official download link\\
PAN & Big-five & 294 users' tweets and their Big-Five personality scores obtained by the BFI-10 questionnaire&\href{https://pan.webis.de/clef15/pan15-web/author-profiling.html}{\scriptsize{https://pan.webis.de/clef15/pan15-web/author-profiling.html}}\\
Kaggle & MBTI & 8,675 users, with each user contributing 45-50 posts & \href{https://www.kaggle.com/datasets/datasnaek/mbti-type}{\scriptsize{https://www.kaggle.com/datasets\break /datasnaek/mbti-type}}\\
Pandora & MBTI & Dozens to hundreds of posts from each of the 9,067 Reddit users & \href{https://psy.takelab.fer.hr/datasets/all/}{\scriptsize{https://psy.takelab.fer.hr/datasets/all/}} \\
FriendsPersona & Big-five & 711 short conversations are extracted and annotated from the first four seasons of Friends TV Show transcripts &\href{https://github.com/emorynlp/personality-detection}{\scriptsize{https://github.com/emorynlp/personality-detection}} \\
CPED & Big-five & 12K dialogues in multi-modal context from 40+ TV shows&\href{https://github.com/scutcyr/CPED}{\scriptsize{https://github.com/scutcyr/CPED}}\\
First Impression & Big-five & 10,000 Youtube video clips of people facing and speaking & \href{https://chalearnlap.cvc.uab.cat/dataset/24/description/}{\scriptsize{https://chalearnlap.cvc.uab.cat/dataset/\break 24/description/}}\\

\cite{peters2024large} &Big-five & Dialogues of 566 participants with a chatbot built on the ChatGPT &\href{https://osf.io/edn3g/.}{\scriptsize{https://osf.io/edn3g/}}\\

\cite{cao2024large} & Big-five &  Demographics of 11,341 public figures from Pantheon 1.0 with manually rated personality traits. & \href{https://osf.io/854w2/}{\scriptsize{https://osf.io/854w2/}}\\
\bottomrule
\end{tabular}
\end{center}
\caption{Open source datasets for personality recognition in LLM}
\label{tab:recognition_resources}
\end{table*}

\end{document}